\documentclass[10pt,twocolumn,letterpaper]{article}

\usepackage[pagenumbers]{cvpr} % To force page numbers, e.g. for an arXiv version

\definecolor{cvprblue}{rgb}{0.21,0.49,0.74}

\usepackage[pagebackref,breaklinks,colorlinks,allcolors=cvprblue]{hyperref}

\usepackage{colortbl}
\usepackage{booktabs}
\usepackage{makecell}
\usepackage{multicol}
\usepackage{multirow}
\usepackage{wrapfig}
\usepackage{caption}
\usepackage{algorithmic}
\usepackage{algorithm}
\usepackage[accsupp]{axessibility}  % Improves PDF readability for those with disabilities.

\newcommand{\methodname}{FlowMotion\xspace}

\definecolor{mygreen}{RGB}{34,139,34}
\definecolor{mygray}{gray}{0.6}
\definecolor{mygray-bg}{gray}{0.9}

\makeatletter
\def\blfootnote{\xdef\@thefnmark{}\@footnotetext}
\makeatother
\def\paperID{*****} % *** Enter the Paper ID here
\def\confName{CVPR}
\def\confYear{2026}

\title{FlowMotion: Training-Free Flow Guidance for Video Motion Transfer}

\author{Zhen Wang$^{1}$, \; Youcan Xu$^{2}$, \; Jun Xiao$^{2}$, \;  Long Chen$^{1, \dagger }$ \\
    \small	$^1$ The Hong Kong University of Science and Technology \quad $^2$State key lab of CAD\&CG, Zhejiang University \\
	{\tt\small zhenwang@ust.hk, youcan@zju.edu.cn, junx@cs.zju.edu.cn, longchen@ust.hk}\\
    { \tt\small {https://github.com/HKUST-LongGroup/FlowMotion}}
    \vspace{-1em}
}

\begin{document}

\twocolumn[{%

\maketitle

\begin{center}
    \centering
    \captionsetup{type=figure}
    \includegraphics[width=1\textwidth]{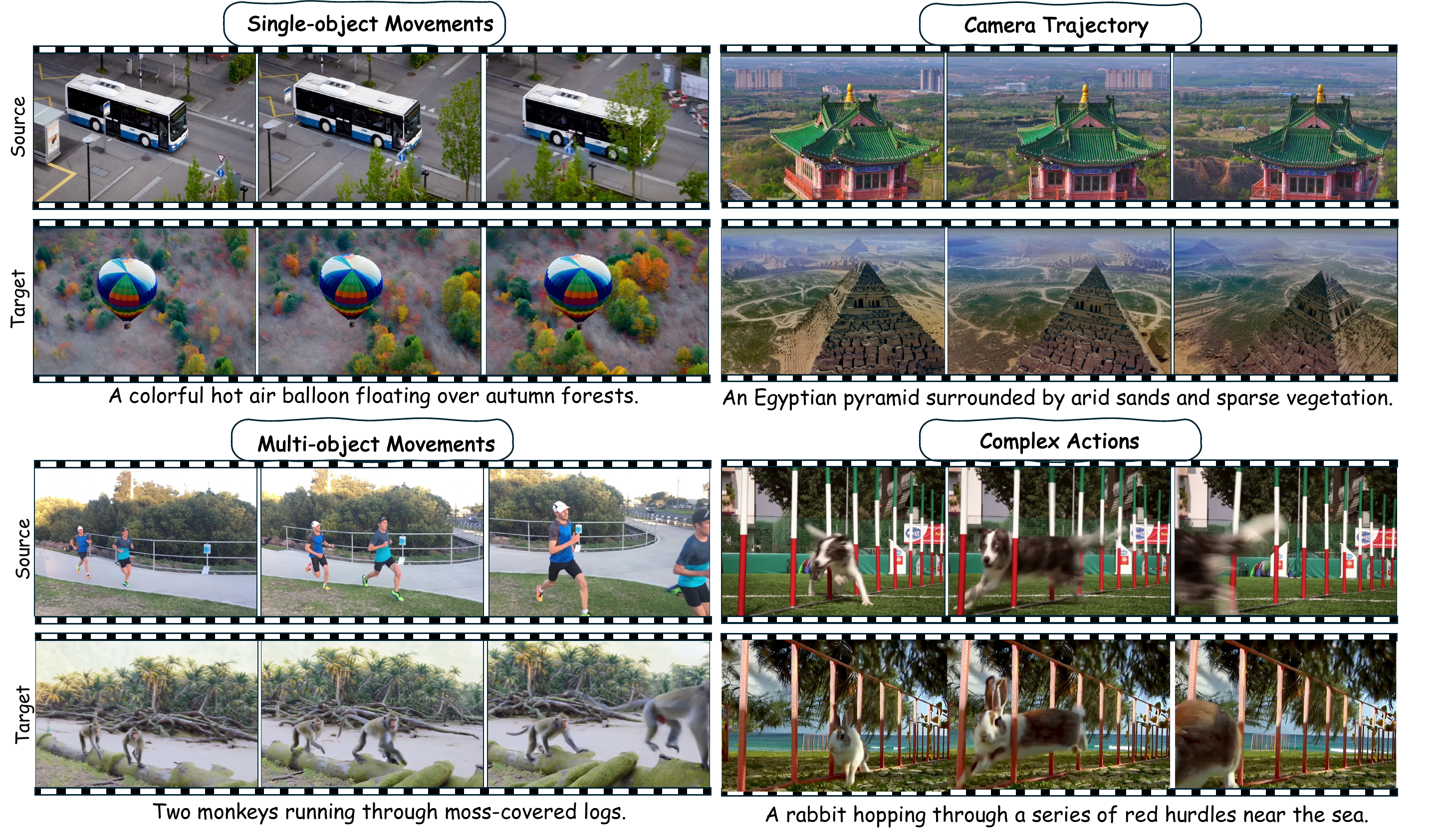}
    \vspace{-2em}
    \caption{\small \textbf{Video Motion Transfer with FlowMotion. } It enables time and resource efficient transfer for various motion types including single- and multi-object movements, camera trajectories and complex actions.} 

    \label{fig:main}
\end{center}%
}]

\begin{abstract}
Video motion transfer aims to generate a target video that inherits motion patterns from a source video while rendering new scenes. Existing training-free approaches focus on constructing motion guidance based on the intermediate outputs of pre-trained T2V models, which results in heavy computational overhead and limited flexibility. In this paper, we present \textbf{FlowMotion}, a novel training-free framework that enables efficient and flexible motion transfer by directly leveraging the predicted outputs of flow-based T2V models. Our key insight is that early latent predictions inherently encode rich temporal information. Motivated by this, we propose flow guidance, which extracts motion representations based on latent predictions to align motion patterns between source and generated videos. We further introduce a velocity regularization strategy to stabilize optimization and ensure smooth motion evolution. By operating purely on model predictions, FlowMotion achieves superior time and resource efficiency as well as competitive performance compared with state-of-the-art methods.
\blfootnote{$^\dagger$Long Chen is the corresponding author.}
\end{abstract}

\section{Introduction}
\label{sec:intro}

Recent advances in large-scale pre-trained Text-to-Video (T2V) models~\citep{yang2024cogvideox, Wan, sora} have demonstrated remarkable capabilities in synthesizing realistic and dynamic video content from textual descriptions. Building upon these generative breakthroughs, a novel and promising direction --- Video Motion Transfer~\cite{yatim2024space,zhao2024motiondirector,pondaven2025video,shi2025decouple} --- has attracted growing attention for customizing and controlling video motions during generation. As illustrated in Figure~1, given source videos, this task aims to transfer realistic motion patterns (\eg, complex object movements, and camera trajectories) to synthesized target videos by pre-trained T2V models, while allowing flexible rendering with diverse new subjects and scenarios. By enabling fine-grained customization and controllability for motion synthesis, video motion transfer has shown great potential for various applications such as virtual reality, filmmaking, and digital entertainment.

The key challenges of this task lie in effectively extracting motion patterns from the source video while ensuring accurate transfer. To achieve this, existing approaches can be broadly categorized into two groups: \textbf{1) Training-based Methods}~\cite{jeong2024vmc,ren2024customize,materzynska2024newmove,wei2024dreamvideo,zhao2024motiondirector,wang2025motioninversion,ma2025follow,shi2025decouple,zhu2025motionrag}: They learn motion representations by fine-tuning pre-trained T2V models on the source video, where a set of parameters (\eg, temporal attention modules~\cite{jeong2024vmc,zhao2024motiondirector,wang2025motioninversion} or LoRAs~\cite{ma2025follow,abdal2025dynamic_concept}) are optimized to capture the motion dynamics. Despite achieving high motion fidelity, these methods require time-consuming training for every reference video, which limits their practicality for real-time or large-scale scenarios. \textbf{2) Training-free Methods}~\cite{yesiltepe2024motionshop,ling2024motionclone,yatim2024space,xiao2024video,pondaven2025video}: To enable efficient transfer without per-video training, another line of research focuses on extracting motion patterns during inference to guide the generation process. These methods analyze the intermediate outputs of pre-trained T2V models, such as temporal attention maps~\cite{ling2024motionclone}, diffusion features~\cite{yatim2024space,xiao2024video}, and cross-frame attention flows~\cite{pondaven2025video} to construct motion representations that serve as guidance signals. The target video is then generated by optimizing its latent representation across denoising steps, minimizing the discrepancy between the motion representations of the source and generated video. By leveraging such \emph{motion guidance}, the transfer can be achieved without updating model parameters.

\begin{figure}[t]
    \centering
    \includegraphics[width=0.475\textwidth]{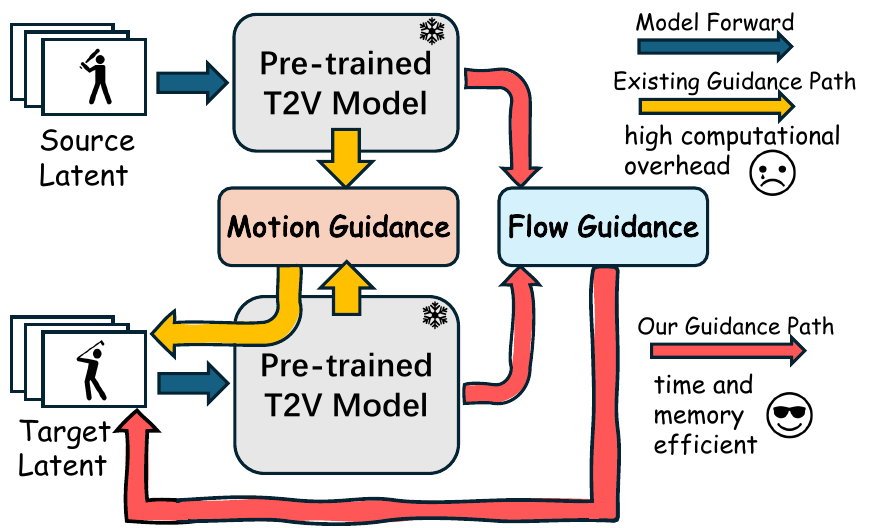}
    \vspace{-1em}
    \caption{Guidance process of exisiting methods and FlowMotion.}
    \vspace{-1em}
    \label{fig:intro}
\end{figure}

While avoiding explicit parameter tuning, existing training-free approaches still suffer from significant \textbf{computational overhead}. As illustrated in Figure~\ref{fig:intro}, current motion guidance mechanisms rely on the intermediate outputs of pre-trained T2V models, \ie, specific feature layers~\cite{yatim2024space,xiao2024video} or attention blocks~\cite{ling2024motionclone,pondaven2025video} within certain architectures such as U-Net~\cite{uNET} or DiT~\cite{Scalable_Diffusion_tf}. Such dependence incurs \emph{substantial memory requirements}~\cite{ling2024motionclone,pondaven2025video} due to gradient backpropagation through internal deep model layers. In addition, some extra computations such as the iterative inversion process~\cite{yatim2024space,xiao2024video} for reference videos and pairwise attention operations~\cite{pondaven2025video} across latent patches and frames further introduce notable \emph{time costs}. These limitations collectively hinder their scalability and generalization ability in real-world video generation scenarios.

To address these limitations, we propose \textbf{FlowMotion}, a novel training-free framework that enables efficient and flexible video motion transfer with \emph{flow guidance}. As shown in Figure~\ref{fig:intro}, the flow guidance directly leverages the predicted outputs of flow-based T2V models to capture and propagate motion patterns in an inversion-free manner, eliminating the need for intermediate outputs or architecture-specific dependencies.

Our method built upon the key observation that the early latent predictions of flow-based T2V models inherently encode rich temporal information with fewer appearance details during video generation. Motivated by this, we propose latent-based \textbf{\emph{flow guidance}}, which extracts and aligns motion representations directly based on the latent predictions of both source and generated videos to guide motion transfer. Specifically, the flow guidance consists of two objectives: it directly aligns latent predictions to maintain global motion consistency, and aligns their frame-wise differences to emphasize temporal variations while suppressing static appearance cues. To further stabilize optimization, we introduce a \textbf{\emph{velocity regularization}} strategy that constrains each velocity update to follow the accumulated flow direction, mitigating over-alignment and ensuring smooth motion evolution. Notably, by operating purely at the model prediction ouput, FlowMotion enables both time- and resource-efficient motion transfer, avoiding architectural dependency and eliminating costly gradient propagation through internal model layers.

Our contributions are threefold: 1) We propose FlowMotion, a novel training-free framework for motion transfer that operates directly on the prediction outputs of pre-trained flow-based T2V models. 2) We provide a comprehensive analysis of flow-based T2V generation, which offer new insights into their behaviors and motivate an efficient and stable motion transfer. 3) Extensive experiments demonstrate that FlowMotion achieves promising performance and efficiency compared with existing state-of-the-art methods, and can further generalize to new backbones.

\section{Related Work}
\label{sec:related_work}

\noindent\textbf{Text-to-Video Generation.}
Text-to-video (T2V) generation has advanced rapidly in recent years, primarily driven by the fast evolution of diffusion models~\citep{score_match,ddpm,nichol2021improved,diffusion_beat_gan,wang2025recent}. Early approaches~\citep{ho2022video,wang2023modelscope,chen2024videocrafter2,guo2023animatediff,blattmann2023stable} typically build upon U-Net~\citep{uNET} architectures, where temporal modules are appended to spatial diffusion backbones to capture video dynamics. However, the limited scalability and locality of U-Net structures constrain their ability to model long-range temporal dependencies and complex motion interactions.
Recent research has embraced Diffusion Transformers (DiTs)~\citep{Scalable_Diffusion_tf}, marking a significant architectural shift in T2V generation. Benefiting from the global receptive field and superior scalability of transformer architectures, DiT-based models~\cite{lu2023vdt,chen2024gentron,gupta2024photorealistic,ma2024latte} such as Sora~\citep{sora} and CogVideoX~\citep{yang2024cogvideox} have achieved remarkable success, producing high-quality and temporally coherent videos from complex textual descriptions.
More recently, the emergence of flow-based generative models~\cite{lipman2022flow,liu2022flow,albergo2022building,li2026path,chen2026bi} has further advanced this field. By integrating flow matching with DiT architectures, models like Wan~\citep{Wan} and Hunyuan Video~\cite{kong2024hunyuanvideo} have demonstrated state-of-the-art performance, enabling more stable training and improved video fidelity, highlighting the growing potential of flow-based T2V models for more efficient and controllable generation tasks.

\noindent\textbf{Video Motion Transfer.}
Video motion transfer aims to reproduce the motion dynamics of a reference video within generated target videos. Existing approaches can be broadly categorized into two types. Training-based methods\cite{jeong2024vmc,ren2024customize,ma2025follow,shi2025decouple} focus on learning motions through model fine-tuning. Early methods build upon U-Net-based~\cite{uNET} architectures, leveraging the inherent separated spatial-temporal attention modules. These approaches typically fine-tune temporal-attention layers~\cite{jeong2024vmc,ruiz2023dreambooth,ren2024customize,materzynska2024newmove,wei2024dreamvideo,zhao2024motiondirector,wang2025motioninversion} while freezing spatial attention layers to preserve appearance-motion decoupling. Recent methods have shifted to handle the more complex 3D full-attention mechanisms~\cite{Scalable_Diffusion_tf} in DiT based models~\cite{yang2024cogvideox,Wan,sora}, employing strategies such as two-stage LoRA training~\cite{ma2025follow,abdal2025dynamic_concept,xu2025compositional} or analyzing specific DiT features~\cite{shi2025decouple}  to better disentangle and learn motion patterns. Although these methods achieve impressive motion fidelity, they require costly training. Training-free methods~\cite{yesiltepe2024motionshop,ling2024motionclone,yatim2024space,xiao2024video,pondaven2025video} extract motion cues from pre-trained T2V models at inference. Such cues are derived from temporal attention maps~\cite{ling2024motionclone}, diffusion features~\cite{yatim2024space,xiao2024video}, or cross-frame attention flows~\cite{pondaven2025video}, and are used to guide video generation by matching motion representations between reference and generated latents through iterative optimization. While training-free and avoid parameter tuning, these methods still incur high memory and time overhead due to costly gradient propagation through internal model layers, which further limits their flexibility and generality.

\section{Method}

\subsection{Preliminary: Latent Space Flow Matching}
\label{method:pre}

Building upon the success of latent diffusion models~\cite{rombach2022high}, state-of-the-art flow matching models (FMs)~\cite{flux2024,esser2024scaling,Wan} perform generative modeling directly in the latent space rather than the pixel space. Typically, a latent FM consists of a pretrained autoencoder and a velocity prediction network. Given input data $x$, the encoder $\mathcal{E}$ maps it into the latent space $z_0 = \mathcal{E}(x)$, where the FM learns a transformation from a standard Gaussian distribution $z_1 \sim \mathcal{N}(0, \mathbf{I})$ to the target $z_0$, and decodes it back to clean data with the decoder $x = \mathcal{D}(z_0)$. Specifically, for each training step with timestep $t\sim \left [ 0, 1 \right ]$, FM obtains a noised intermediate latent via the linear interpolating between $z_0$ and $z_1$ as $z_t = (1-t) z_0 + t z_1 $. The velocity prediction network $v_\theta$ is then trained to estimate the velocity $v_t = \frac{dz_t}{dt} = z_1 - z_0$ by minimizing the flow matching loss:
\begin{equation}
\label{eq:fm_training_objective}
\mathcal{L}_\mathrm{FM}=\mathbb{E}_{z_0\sim\mathcal{E}(x),z_1\sim\mathcal{N}(0,1),t}\left[\|v_\theta(z_t,t)- (z_1 - z_0)\|_2^2 \right].
\end{equation}
During inference, the FM starts from a random noisy latent $z_1 \sim \mathcal{N}(0, \mathbf{I})$, and progressively integrates the predicted velocity to obtain the clean latent: $z_0 = z_1 - \int_{1}^{0}v_\theta(z_t;t)dt$.

\subsection{Motivation}
\label{sec:motivation}
Formally, given a source video $\mathcal{V}$ and a text prompt $P$, the task of video motion transfer aims to synthesize a target video $\mathcal{J}$ that preserves the motion patterns from $\mathcal{V}$ while adapting its appearance and content to the semantics of $P$. Existing training-free approaches extract motion representations from the intermediate layers or attention blocks of pre-trained T2V models and use them as guidance signals. In contrast, we shift our focus from intermediate features to the model’s predicted outputs, exploring whether the intrinsic predictions of flow-based T2V models contain sufficient motion information for direct and efficient transfer.

\begin{figure}[t]
    \centering
    \includegraphics[width=0.475\textwidth]{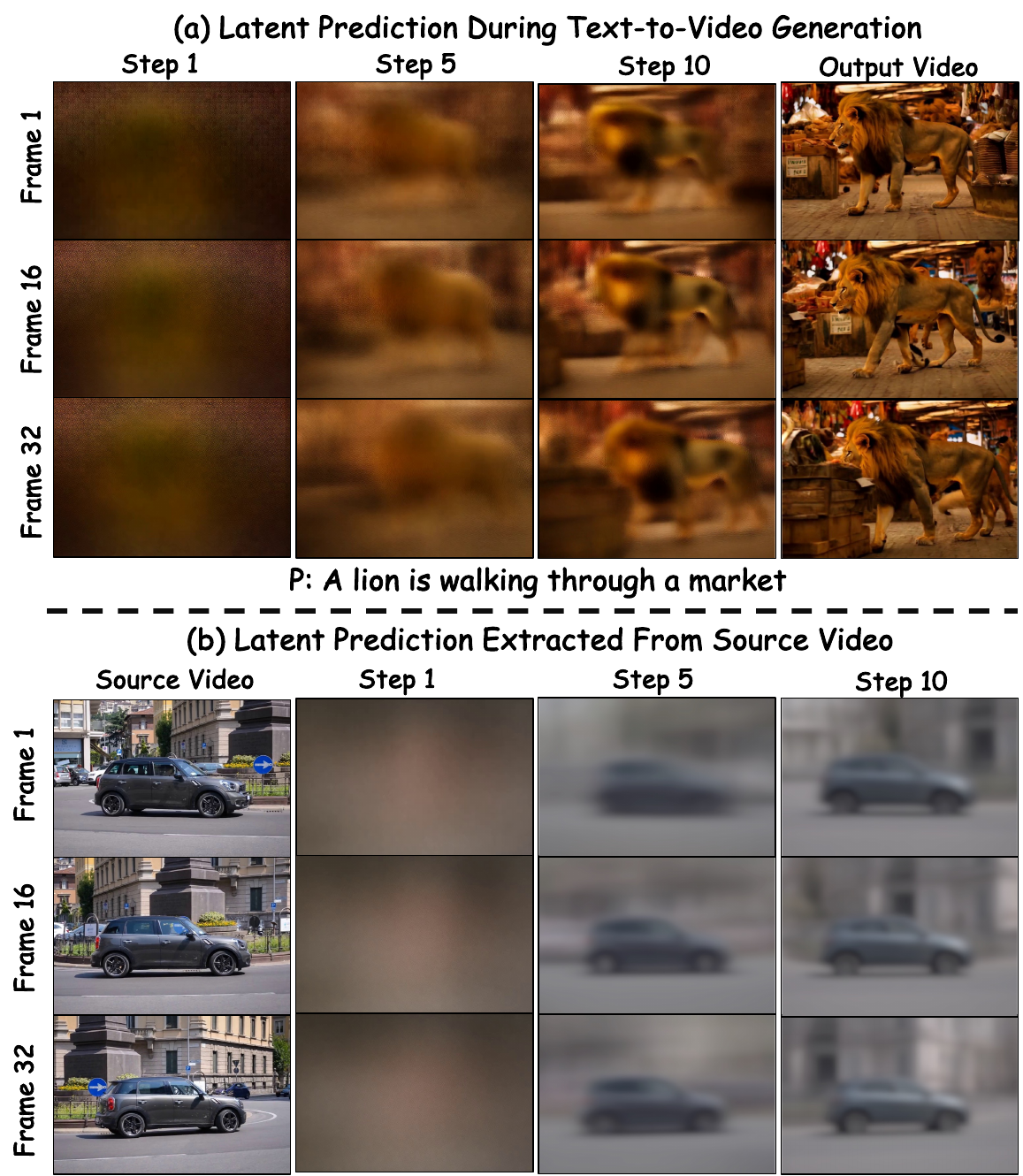}
    \vspace{-2em}
    \caption{The visualization of latent prediciton: (a) During T2V generation. (b) Extracted from source video.}
    \vspace{-1em}
    \label{fig:motivation}
\end{figure}

\noindent\textbf{Analysis of Flow-based T2V Generation.}
To verify this hypothesis, we analyze the generation behavior of flow-based T2V models, taking Wan~\cite{Wan} as a representative example. As described in Sec.~\ref{method:pre}, the model starts from a noisy latent, and at each denoising step, outputs an instantaneous velocity prediction conditioned on the prompt as $v_t = v_\theta(z_t, t, P)$, which represents the directional change of the latent, guiding the transformation from the noise distribution toward the clean latent space. In the standard denoising process, $v_t$ is used to make small-step updates of $z_t$, gradually removing noise. However, both $z_t$ and $v_t$ remain noisy during generation, making it difficult to directly examine how the video information evolve over time. To better interpret the generation process, we introduce a \emph{latent prediction} for each step, estimating the clean latent in a single projection:
\begin{equation}
\label{eq:latent_prediction}
\hat{z}_0(t) = z_t - t\, v_t.
\end{equation}
Hence, $\hat{z}_0(t)$ can be viewed as a one-step approximation of the final clean latent that the model’s current velocity points toward. By decoding $\hat{z}_0(t)$ through the pretrained decoder, we obtain intermediate visualizations $\mathcal{D}(\hat{z}_0(t))$ that reveal the evolution of appearance and motion during denoising. As shown in Figure~\ref{fig:motivation}(a), for a standard 50-step T2V generation, coarse temporal information begin to emerge from the very first step, where the overall location and motion direction become roughly visible. Within about 5 steps, the object motion trajectory and camera movement (\eg, moving from right to left) appears, and by around 10 steps, the fine-grained actions (\eg, limb movements) and entire scene dynamics become clear. As denoising proceeds, finer appearance cues gradually accumulate, refining object textures and scene precision. These observations indicate that the early-stage latent predictions inherently encode rich temporal information --- evolving progressively from coarse location to object trajectories, and finally the fine-grained actions and scene-level dynamics.

\noindent\textbf{Motion Representation Extraction.}
Motivated by the above observations, we regard the early-stage latent predictions as a natural representation of motion. We then proceed to extract such motion representations from the source video for guidance. Concretely, we first encode the source video into the clean latent $z^{src}_0 = \mathcal{E}(\mathcal{V})$. Instead of performing time-consuming inversion, we directly obtain the noisy source latent $z^{src}_t$ via forward noising, where at each step $t$, a random noise latent $\epsilon_t \sim \mathcal{N}(0, \mathbf{I})$ is sampled and used for linear interpolation $z^{src}_t = (1-t) z^{src}_0 + t \epsilon_t $. This noisy latent is then fed into the flow-based T2V model with an empty prompt to predict the instantaneous velocity $v^{src}_t = v_\theta(z^{src}_t, t, \emptyset )$, from which we estimate the corresponding latent prediction for the source video as:
\begin{equation}
\label{eq:latent_prediction_src}
\hat{z}^{src}_0(t) = z^{src}_t - t v^{src}_t.
\end{equation}
As visualized in Figure~\ref{fig:motivation}(b), the latent predictions from the source video exhibit similar progressive evolutions as observed in the generation process --- capturing temporal dynamics that unfold from coarse trajectories (\eg, from left to right) to detailed object actions (\eg, car turning) and scene transitions (\eg, background shifts). This inversion-free extraction provides an efficient way to obtain accurate and interpretable motion representations directly from the flow-based latent predictions, reflecting the intrinsic motion cues embedded in the model’s generative flow.

\begin{figure*}[t]
    \centering
    \includegraphics[width=0.95\textwidth]{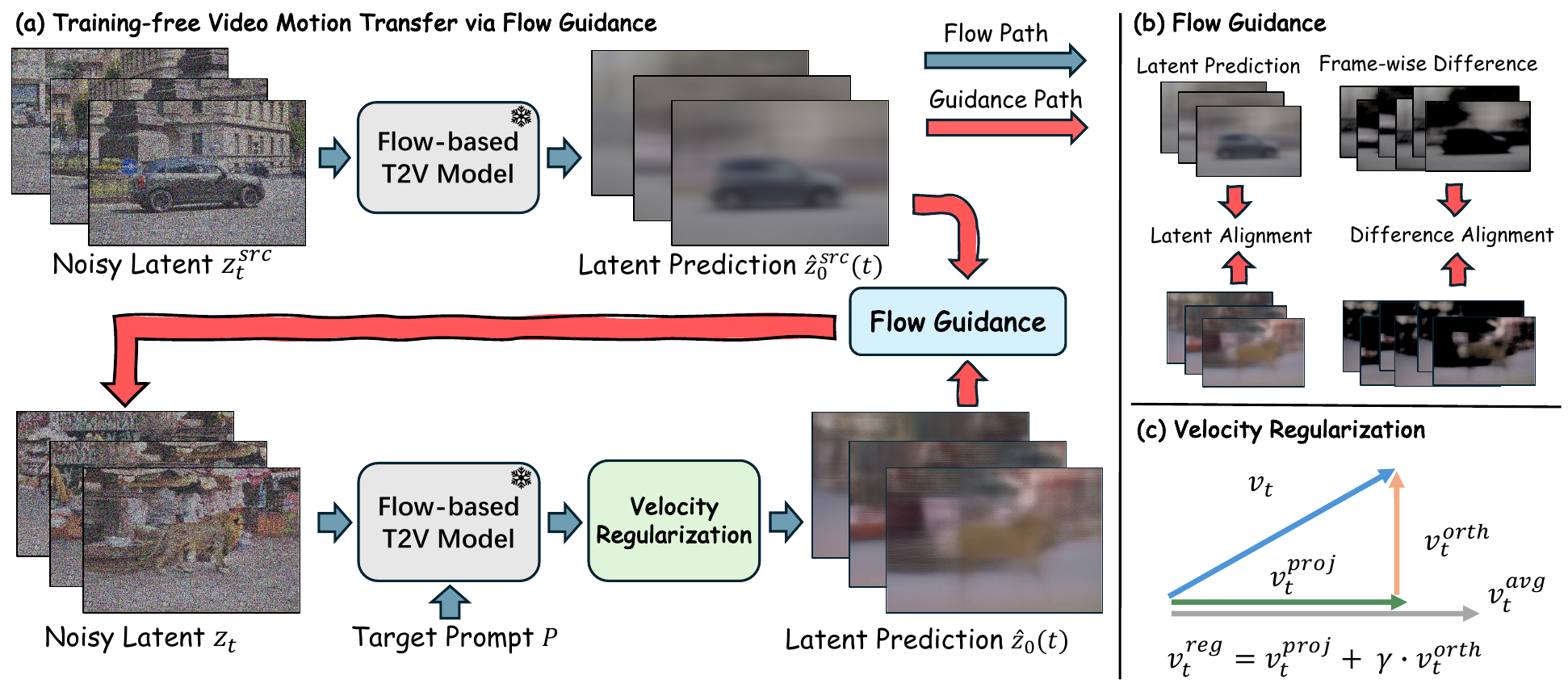}
    \vspace{-1em}
    \caption{\textbf{The overview of FlowMotion.} (a) Training-free video motion transfer with flow guidance and velocity regularization. (b) Two objective of flow guidance. (c) The velocity regularization process.}
    \vspace{-1em}
    \label{fig:pipeline}
\end{figure*}

\subsection{Flow Guidance}
Building upon the above insights, we propose a latent-based flow guidance to explicitly steer the motion pattern in the target video, and the core idea is to align the latent predictions of the target video with those of the source video. As shown in Figure~\ref{fig:pipeline}(a), for a denoising step with timestep $t$, we first obtain the latent predictions of the source and target videos $\hat{z}^{src}_0(t), \hat{z}_0(t) \in \mathbb{R}^{F\times H\times W\times C}$ following the process described in Sec.~\ref{sec:motivation}, where $F$, $H$, $W$, and $C$ representing the frame number, height, width, and channels, respectively.

As illustrated in Figure~\ref{fig:pipeline}(b), the flow guidance consists of two objectives:  
1) Latent Alignment (LA): directly aligning the source and target latent predictions to ensure overall motion consistency. 2) Difference Alignment (DA): emphasizing temporal variations while suppressing static appearance information by computing the frame-wise latent differences $\bigtriangleup(\hat{z}^{src}_0(t))$ and $\bigtriangleup(\hat{z}_0(t)) \in \mathbb{R}^{F\times(F-1)\times H\times W\times C}$, and aligning them to preserve dynamic motion cues. The overall flow guidance loss is then formulated as:
\begin{align}
\label{eq:loss}
 &\mathcal{L}_{FG}  = \alpha \cdot \mathcal{L}_{LA} + \beta \cdot \mathcal{L}_{DA} \\
 &= \alpha \left \| \hat{z}^{src}_0(t)- \hat{z}_0(t) \right \|^{2}_{2}
 + \beta \left \| \bigtriangleup (\hat{z}^{src}_0(t)) - \bigtriangleup (\hat{z}_0(t)) \right \|^{2}_{2} . 
\end{align}
where $\alpha$ and $\beta$ are the weighting coefficients controlling the balance between global motion alignment and fine-grained temporal dynamics. The target latent $z_t$ is then optimized by backpropagating $\mathcal{L}_{FG}$, effectively guiding the generative flow to follow the motion pattern of the source video. Notably, this latent-based guidance operates directly on the prediction outputs, eliminating the need for gradient propagation through internal layers and thus avoiding substantial computational overhead.

\subsection{Velocity Regularization for Stable Guidance}
Although the proposed flow guidance efficiently aligns motion patterns, directly optimizing through latent predictions may lead to over-alignment with appearance details (\eg, object shapes) and cause unstable updates across timesteps, resulting in degraded overall quality and temporal consistency. To address these issues, we introduce a velocity regularization strategy that stabilizes the generative flow during iterative  optimization. Specifically, at each denoising step with timestep $t$, we compute an \emph{average velocity} that represents the overall flow direction accumulated so far:
\begin{equation}
\label{eq:avg}
v^{avg}_{t} = \frac{(z_t - z_1)}{(t-1)}.
\end{equation}
As shown in Figure~\ref{fig:pipeline}(c), we then decompose the current velocity $v_t$ into a projection component $v^{proj}_t$ along $v^{avg}_t$ and an orthogonal component $v^{orth}_t$:
\begin{align}
\label{eq:proj}
v^{proj}_{t} = \frac{\langle v_t, v^{avg}_t \rangle}{\|v^{avg}_t\|_2^2} v^{avg}_t, \quad 
v^{orth}_t = v_t - v^{proj}_t.
\end{align}
To avoid abrupt directional shifts, we decay the orthogonal component with a factor $\gamma \in [0,1]$:
\begin{equation}
\label{eq:reg}
v^{reg}_{t} = v^{proj}_{t} + \gamma \cdot v^{orth}_{t}.
\end{equation}
The regulated velocity $v^{reg}_{t}$ is then used to predict the latent:
\begin{equation}
\label{eq:latent_prediction2}
\hat{z}_0(t) = z_t - t\, v^{reg}_{t}.
\end{equation}
Intuitively, this velocity regularization strategy constrains each update to follow the accumulated flow direction while suppressing excessive deviations. By doing so, it mitigates over-alignment and ensures more stable motion evolution across optimization steps, thereby preserving overall visual quality and temporal coherence.

\section{Experiment}

\begin{figure*}[t]
    \centering
    \includegraphics[width=1\textwidth]{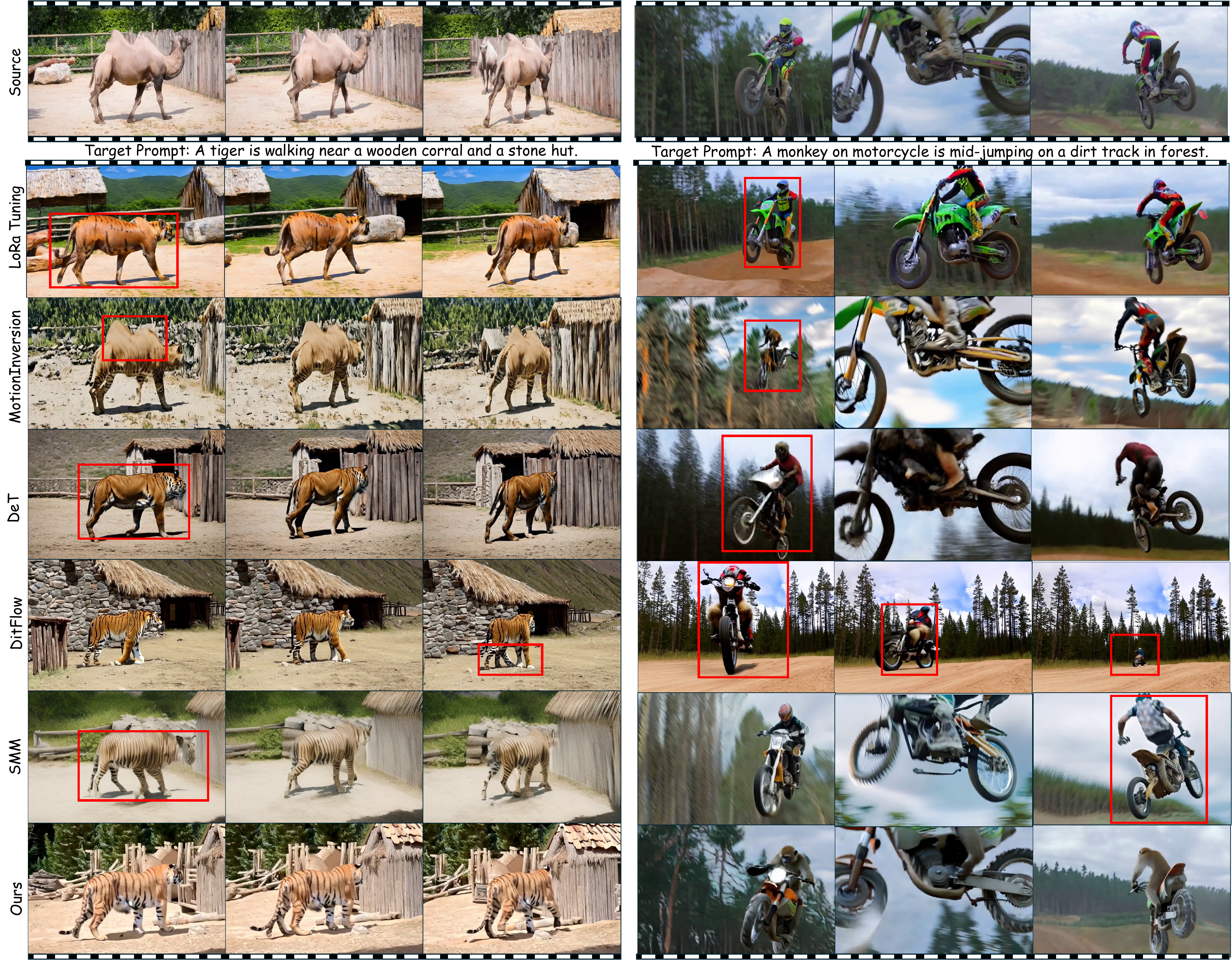}
    \vspace{-2em}
    \caption{\textbf{Qualitative comparison with SOTA methods.} \textcolor{red}{Red boxes} indicated low quality content across frames.}
    \vspace{-1em}
    \label{fig:qualitative_comparison}
\end{figure*}

\subsection{Experimental Setup}
\noindent\textbf{Implementation Details.} We apply FlowMotion on two pre-trained flow-based T2V models --- Wan2.1 and Wan2.2~\cite{Wan}. Specifically, we employ 50 denoising steps for T2V generation with a classifier-free guidance of 6, and apply the optimization for the initial 10 steps. Generally, for Wan2.1, we use the Adam~\cite{kingma2014adam} optimizer with a learning rate of 0.003 for 3 optimization steps. We set $\alpha:\beta = 4:1$ for the loss, and $\gamma = 0.1$ for velocity regularization\footnote{\noindent~Due to limited space, more details and settings are in the Appendix. \label{ft:appendix}}.

\noindent\textbf{Dataset and Metrics.} We collected 50 high-quality videos from MTBench~\cite{shi2025decouple} and previous open sources~\cite{pont20172017}, covering diverse motion types and object domains\footref{ft:appendix}. Each video has a resolution of $480\times720$ with 49 frames and is paired with three target prompts. For {quantitative evaluation}, we adopt three standard metrics following prior works~\citep{zhao2024motiondirector,shi2025decouple}: (1) {Text Similarity}, computed by the average CLIP score~\citep{CLIP} between generated frames and the target text prompt. (2) {Motion Fidelity}, computed as the holistic similarity between motion tracklets~\citep{yatim2024space} extracted from the source and generated videos. (3) {Temporal Consistency}, computed by the average CLIP feature similarity between consecutive frames. In addition, we evaluate \emph{computational efficiency} in terms of training and inference time, as well as GPU memory requirement per video.

\noindent\textbf{Baselines.} We compare FlowMotion against state-of-the-art methods, including training-based methods: MotionDirector~\cite{zhao2024motiondirector}, MotionInversion~\cite{wang2025motioninversion}, DeT~\cite{shi2025decouple} and LoRA~\cite{hu2022lora}. Training-free methods: MotionClone~\cite{ling2024motionclone}, MOFT~\cite{xiao2024video}, SMM~\cite{yatim2024space} and DiTFlow~\cite{pondaven2025video}. Ideally, all baselines should be implemented on the same backbone for a fair comparison. However, we found it infeasible in practice: (1) U-Net–based methods are incompatible with recent DiT-based backbones, and (2) their architecture-dependent designs cause severe performance degradation when directly switching backbones. Therefore, to ensure both fairness and validity, we adopt each baseline’s official backbone and implementation, allowing them to fully demonstrate designed capabilities. Meanwhile, we select backbones with comparable parameter scales to maintain overall fairness. Unless stated otherwise, we adopt the backbones detailed in Table~\ref{tab:num_comparison}. All methods run on NVIDIA H20-95G GPU.

\subsection{Qualitative Results}
\begin{figure*}[t]
    \centering
    \includegraphics[width=1\textwidth]{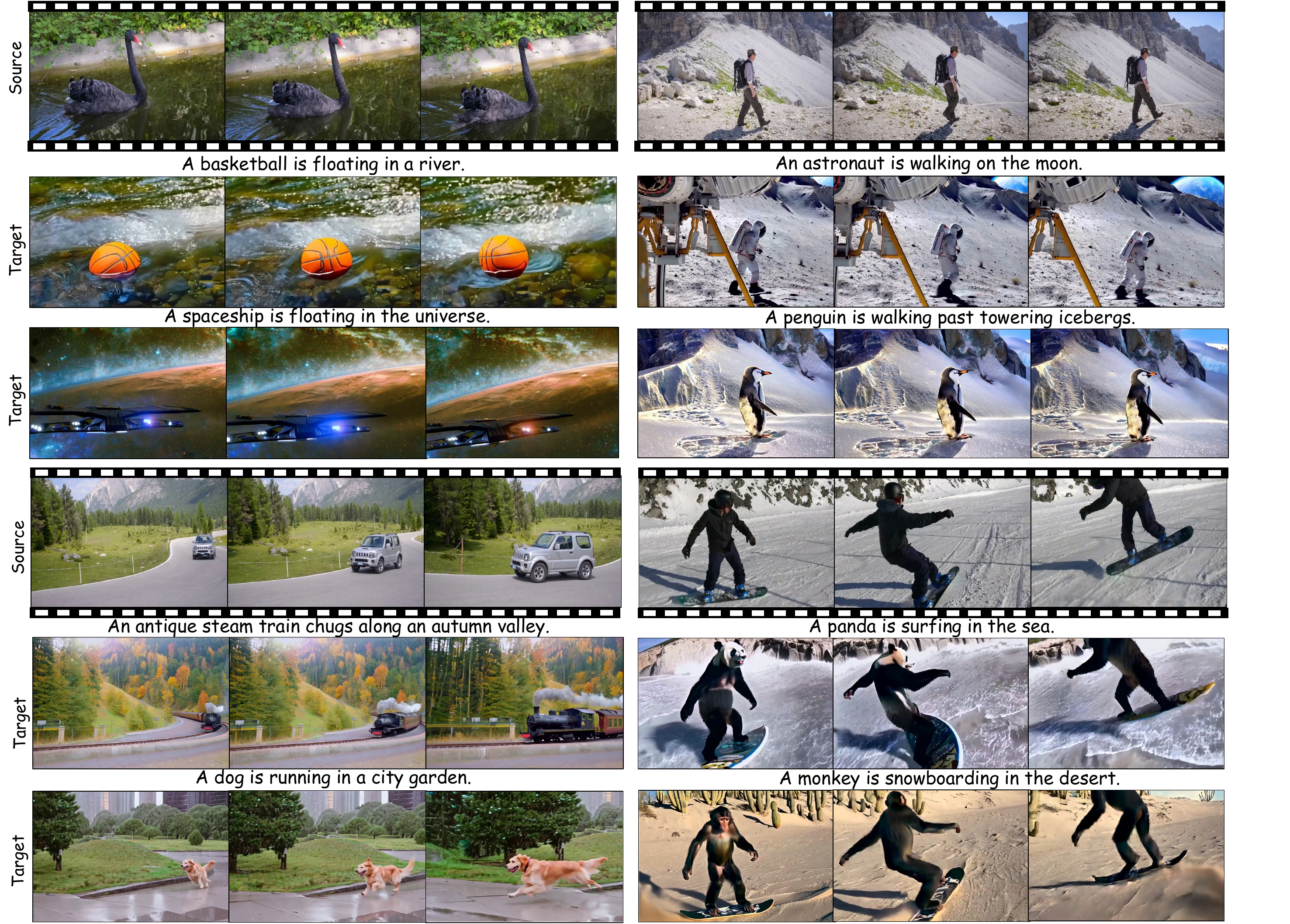}
    \vspace{-2em}
    \caption{\textbf{Qualitative results of FlowMotion.} The left column generated by Wan2.1-1.3B, while right column generated by Wan2.2-5B.}
    \vspace{-1em}
    \label{fig:more_res}
\end{figure*}

As shown in Fig.~\ref{fig:qualitative_comparison}, we observe that: 1) Training-based methods (first three rows), while achieving high motion fidelity, suffer from severe overfitting, \ie, the generated target subjects are strongly influenced by the source video. For example, the generated \texttt{tiger} resembles the \texttt{camel} in the source video, or loses its distinctive characteristics. And they all fail to synthesize the intended subject \texttt{monkey}, instead retaining the human appearance. 2) Training-free methods DitFlow and SMM struggle to preserve motion fidelity, especially in complex scenarios involving multiple objects (\eg, monkey riding a motorcycle) and camera motion changes (\eg, right-to-left panning), resulting in degraded overall visual quality and temporal smoothness. 3) In contrast, FlowMotion achieves a more balanced trade-off, maintaining both high-fidelity motion transfer and strong consistency with the target prompt. We show additional results in Fig~\ref{fig:more_res}, where FlowMotion is applied to different flow-based backbones. It is evident that our method faithfully transfers motion from source videos while enabling flexible text control, \eg, from \texttt{swan} to \texttt{basketball} and \texttt{car} to \texttt{dog}, together with new scenes. Moreover, FlowMotion can also be applied to larger backbones like Wan2.2-5B, producing richer content and complex motions. This further suggests the potential of our method for broader applicability and generalization.

\begin{table*}[t]
  \centering
 % \vspace{-2mm}

  \resizebox{0.95\textwidth}{!}{%
    % \small  % Increased font size
    \begin{tabular}{l|c|c|ccc|ccc}
      \toprule
      \multirow{2}{*}{Method} & \multirow{2}{*}{Backbone} & \multirow{2}{*}{Param}
        & \multicolumn{3}{c|}{Quantitative Metrics}
        & \multicolumn{3}{c}{Computational Efficiency} \\
      \cmidrule(lr){4-6}\cmidrule(lr){7-9}
        & & & Text Sim.$\uparrow$ & Motion Fid.$\uparrow$  & Temp. Cons. & 
        Training Time (s) & Inference Time (s) & GPU Memory (G) \\
      \midrule
      \multicolumn{9}{l}{\textbf{\emph{Training-based}}} \\
      \midrule
    LoRA Tuning~\cite{hu2022lora} & Wan2.1& 1.3B
        & 0.327 & 0.782 & 0.977 & 8100
        & 135  & 25.0 \\
    MotionDirector~\cite{zhao2024motiondirector} & ZeroScope&0.7B
        & 0.335 & 0.801 & 0.969 & 1662
        &  140 & 28.0 \\
    MotionInversion~\cite{wang2025motioninversion} &ZeroScope &0.7B
        & 0.328 & \underline{0.839} & 0.970 & 1170
        & 115 & 24.0\\
    DeT~\cite{shi2025decouple} &CogVideoX&2B
    & 0.340 & {0.812} & 0.980 & 
       2760 & 133 & 20.0 \\
      \midrule
      \multicolumn{9}{l}{\textbf{\emph{Training-free}}} \\
      \midrule
    MotionClone~\cite{ling2024motionclone} &AnimateDiff & 1.3B
        & 0.332 & 0.786  & 0.940 & - 
        &  804 & 51.5 \\
    MOFT~\cite{xiao2024video} &AnimateDiff & 1.3B
        & 0.338  & 0.582  & 0.973 & - 
        &  576 & 75.0 \\
    SMM~\cite{yatim2024space} &ZeroScope& 0.7B
        & 0.322 & 0.762  & 0.958 & - 
        & 1839  & 89.4  \\
    DiTFlow~\cite{pondaven2025video} & CogVideoX & 2B
        & \textbf{0.350} & 0.691  &\underline{0.983} &   -
        & 349  & 63.5 \\
      Ours  & Wan2.1 & 1.3B
        & \cellcolor{mygray-bg}{\underline{0.347}
        }& \cellcolor{mygray-bg}{\textbf{0.850}}  & \cellcolor{mygray-bg}{\textbf{0.986}} &  \cellcolor{mygray-bg}{-}
        & \cellcolor{mygray-bg}{213}  & \cellcolor{mygray-bg}{19.3} \\
      \bottomrule
    \end{tabular}%
  }
  \vspace{-0.5em}
    \caption{\textbf{Quantitative comparison with SOTA video motion transfer methods}. Best results in \textbf{bold} and second best are \underline{underlined}. } 
      \label{tab:num_comparison}
\vspace{-1.5em}
\end{table*}

\subsection{Quantitative Results}

As shown in Table~\ref{tab:num_comparison}, we can observe 1) FlowMotion achieves the highest motion fidelity and temporal consistency among all methods. Although DitFlow obtains the highest text similarity, its motion fidelity is notably poor. MotionInversion suffers from overfitting, results in high motion fidelity but degraded text alignment and temporal consistency. In contrast, FlowMotion achieves the most balanced trade-off, maintaining strong text alignment while preserving accurate motion dynamics. 2) Training-based methods generally yield more stable and better overall performance compared with training-free baselines, but at the cost substantial training time. Meanwhile, although training-free baselines avoid per-video training, their guidance mechanisms still incur high GPU memory usage, \eg, at least twice that of training-based methods. In particular, SMM, despite using a lightweight 0.7B backbone, consumes up to 89.4 GB of memory and requires over 30 minutes of inference due to additional inversion-based designs. 3) By contrast, FlowMotion eliminates the need for inversion and avoids gradient propagation through internal layers, achieving superior time and resource efficiency while maintaining competitive visual quality and motion fidelity.

\begin{table}[t]
    \centering
    \resizebox{0.48\textwidth}{!}{%
    \begin{tabular}{l|c|c|c|c}
    \hline
        Model & Motion.$\uparrow$ & Temp.$\uparrow$ & Text Align.$\uparrow$ & Overall.$\uparrow$ \\ \hline
MotionInversion~\cite{wang2025motioninversion} &3.406 &3.344 &2.690 & 2.828\\
    SMM~\cite{yatim2024space} & 3.157& 3.090
        & 2.851& 2.665\\
        DeT~\cite{shi2025decouple} & 3.874& 3.831 & 3.378 & 3.465  \\
        DiTFlow~\cite{pondaven2025video} & 2.481 & 3.176 & 3.156& 2.626 \\
        Ours &\cellcolor{mygray-bg}{4.514 }& \cellcolor{mygray-bg}{4.523} & \cellcolor{mygray-bg}{4.505} & \cellcolor{mygray-bg}{4.446} \\
        \hline
    \end{tabular}
    }
      \vspace{-1em}
    \caption{{Average score of user study}. } 
\label{tab:us}
    \vspace{-1em}
\end{table}

\begin{table}[t]
    \centering
    \resizebox{0.48\textwidth}{!}{%
    \begin{tabular}{l|c|c|c}
    \hline
         & Text Similarity$\uparrow$ & Motion Fidelity.$\uparrow$  & Temporal Consis. $\uparrow$  \\ \hline
        w/o DA & 0.341 & 0.842 & 0.981 \\
        w/o VR & 0.313 & 0.809 & 0.968\\
        Ours & \cellcolor{mygray-bg}{0.347} & \cellcolor{mygray-bg}{0.850}  & \cellcolor{mygray-bg}{0.986} \\
        \hline
    \end{tabular}
    }
      \vspace{-1em}
    \caption{{Ablation results of different key designs}.} 
    \vspace{-1em}
\label{tab:ab}
\end{table}

\begin{figure}[t]
    \centering
    \includegraphics[width=0.475\textwidth]{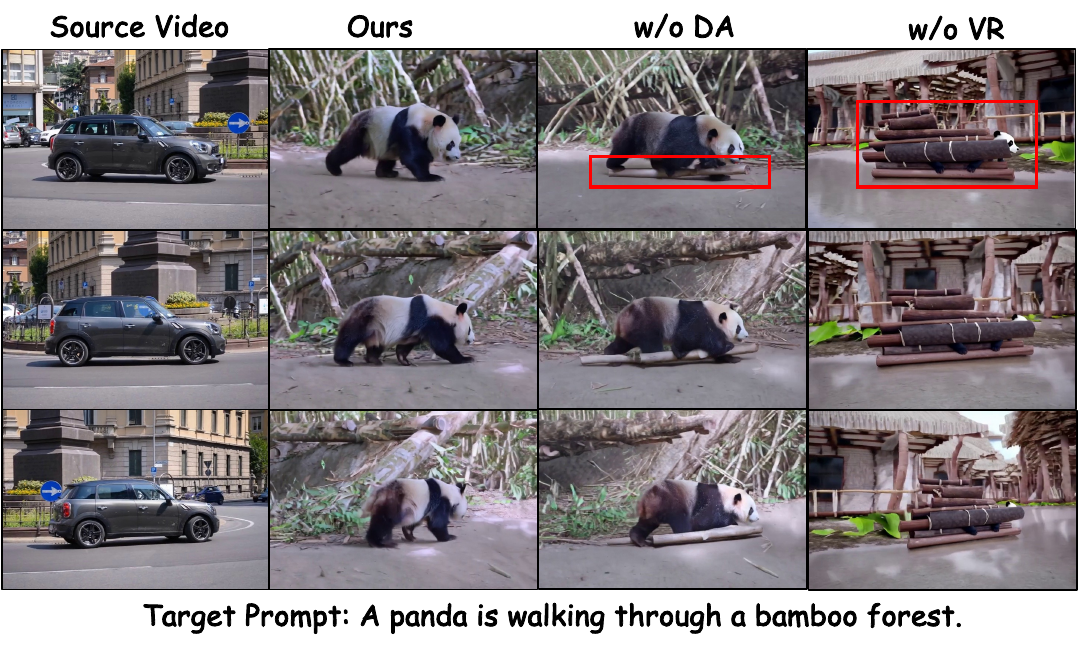}
    \vspace{-2em}
    \caption{The visualization of ablation on key designs. \textcolor{red}{Red boxes} indicate low quality content across frames.}
    \vspace{-1.5em}
    \label{fig:abl}
\end{figure}

\noindent\textbf{User Study.} We further perform human evaluation to reflect real preferences. We invited 20 volunteers to score motion transfer results across four aspects including motion preservation, temporal consistency, text alignment, and overall quality from 5 (best) to 1(worst). As shown in Table~\ref{tab:us}, FlowMotion also achieves the best human preference. 

\subsection{Ablation Study}

\noindent\textbf{Effectiveness of Key Designs.} We first run ablations to verify the design of guidance loss and velocity regularization.

\noindent\emph{Results.} As shown in Table~\ref{tab:ab} and Figure~\ref{fig:abl}, we can observe: 1) Removing the difference alignment (DA) causes performance drop, with noticeable artifacts in appearance (\eg, panda’s feet) and weakens motion coherence, highlighting its role in preserving dynamic motion cues. 2) Excluding the velocity regularization (VR) results in severe quality degradation (\eg, panda mixed with bamboo) due to unstable optimization and overfitting. With all components enabled, FlowMotion achieves the best overall performance. 

\noindent\textbf{Memory Efficiency of Flow Guidance.} While {FlowMotion} achieves significantly lower memory usage than training-free approaches with similar or even larger parameter sizes, we further conduct a comprehensive analysis on the same backbone, \texttt{Wan2.1-1.3B}. Specifically, we simulate to construct guidance from different outputs.

\noindent\emph{Results.} As shown in Table~\ref{fig:memory}, constructing guidance directly from cross-frame attention or attention features leads to out-of-memory, and using velocity outputs as guidance still demands high memory. In contrast, our flow guidance first computes latent predictions from velocity outputs and then employs them for guidance, effectively bypassing gradient propagation through internal layers. This design introduces only negligible memory beyond inference, demonstrating the efficiency and elegance of our approach.

\begin{figure}[t]
    \centering
    \includegraphics[width=0.475\textwidth]{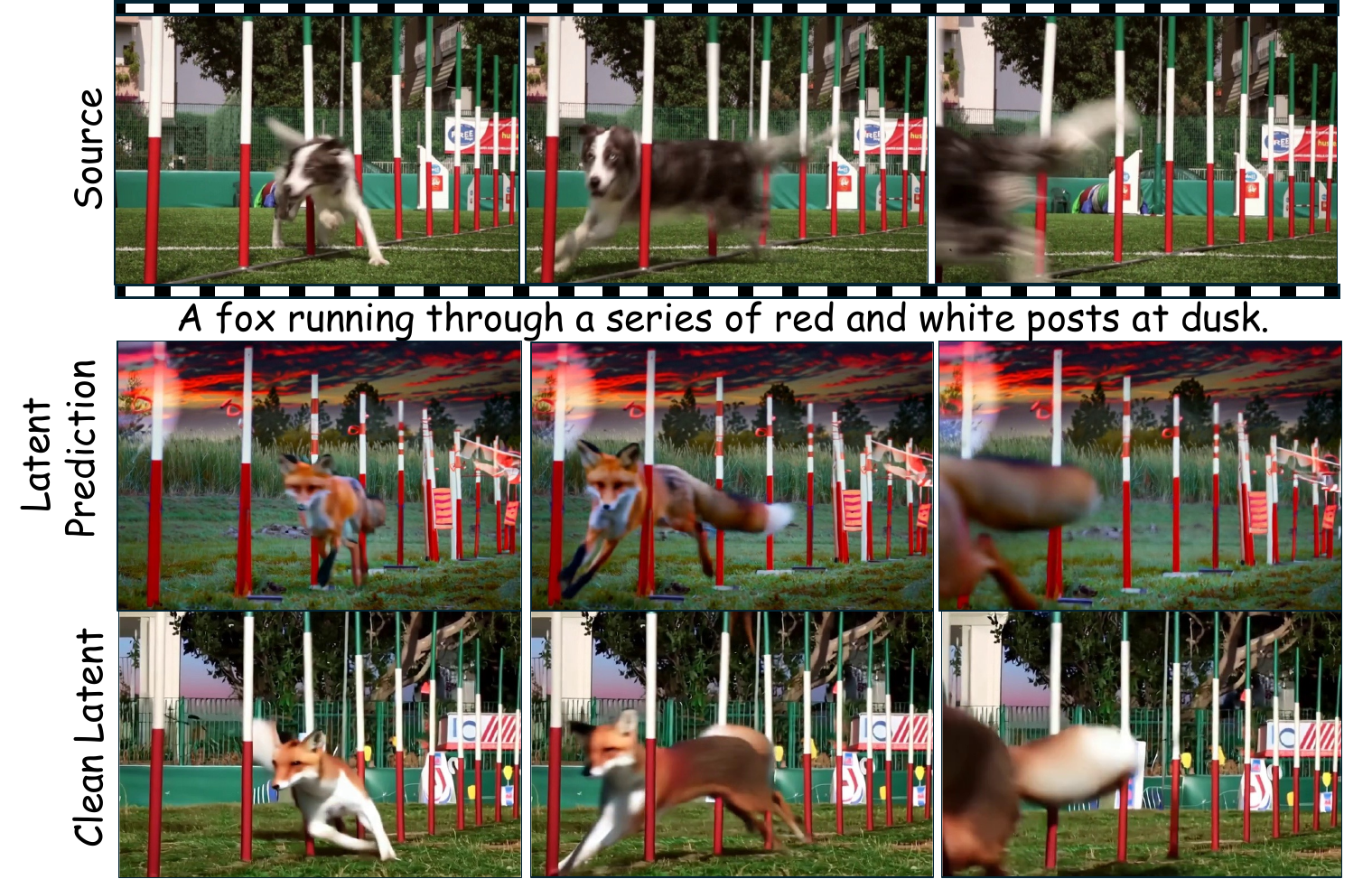}
    \vspace{-2em}
    \caption{Variance of different source motion representation.}
    \vspace{-1.5em}
    \label{fig:variance}
\end{figure}

\noindent\textbf{Variant of Source Motion Representation.}
While early-stage latent predictions are adopted as motion representations due to their rich temporal dynamics, an alternative is to directly use the clean source latent $z^{src}_0$ for guidance, which provides stronger and clearer signals. As shown in the Figure~\ref{fig:variance}, replacing $\hat{z}^{src}_0(t)$ with $z^{src}_0$ enables more precise motion transfer, especially in fine-grained actions, though it may also reduce text alignment and background diversity. In this work, we mainly adopt the latent predictions, while $z^{src}_0$ serves as an optional alternative for complex cases. These results also indicate the potential of developing more adaptive latent-level guidance for jointly improving motion and text control. We provide more discussion in Appendix.

\begin{table}[t]
    \centering
    \resizebox{0.48\textwidth}{!}{%
    \begin{tabular}{l|c|c|c|c}
    \hline
         & Infer & Latent Prediction & Velocity & Att Map\&Feature \\ \hline
        Memory (G) & \textcolor{gray}{17.7} & 19.3 & 93.1 & OOM\\
        \hline
    \end{tabular}
    }
      \vspace{-1em}
    \caption{{Memory requirment of different guidance term}. } 
    \label{fig:memory}
\vspace{-1em}
\end{table}

\section{Conclusion}
In this paper, we introduced FlowMotion, a novel training-free framework for effective and flexible video motion transfer. By directly leveraging the prediction outputs of flow-based T2V models, FlowMotion achieves both time- and resource-efficient motion transfer without relying on intermediate features. Together with our analysis and insights into the flow-based T2V generation, we hope this work can serve as a starting point toward more adaptive, generalizable and controllable video motion transfer frameworks.

\noindent\textbf{Acknowledgment.} This work was supported by the National Natural Science Foundation of China Young Scholar Fund Category B (62522216), Hong Kong SAR RGC General Research Fund (16219025), National Natural Science Foundation of China Young Scholar Fund Category C (62402408), and Hong Kong SAR RGC Early Career Scheme (26208924). This work was also supported the Key R\&D Program of Zhejiang (2025C01128), National Natural Science Foundation of China (62441617), Zhejiang Provincial Natural Science Foundation of China (No.LD25F020001) and Fundamental Research Funds for the Central Universities (226-2025-00057).
% \clearpage
{
    \small
    \bibliographystyle{ieeenat_fullname}
    \bibliography{main}
}
\clearpage


\appendix

\maketitlesupplementary

This supplementary document is organized as follows:
\pagestyle{empty}  % no page number for the second and the later pages
\thispagestyle{empty} % no page number for the first page
\begin{itemize}

    % \item[$\bullet$] We show more video motion transfer results produced by FlowMotion in \textcolor{red}{demo.mp4}.

    \item[$\bullet$]  In Sec.~\ref{sec:motion_rep}, we show more analysis of flow-based T2V model for extracting the motion representation.

    \item[$\bullet$]  In Sec.~\ref{sec:memory}, we provide more discussion and justification of the memory efficiency of FlowMotion.

    \item[$\bullet$]  In Sec.~\ref{sec:variant}, we provide more discussion about the variant of source motion representation.

    \item[$\bullet$]  In Sec.~\ref{sec:algri}, we show the inference scheme.
        
    \item[$\bullet$]  In Sec.~\ref{sec:experimental}, we show more details about the experimental setup.

    \item[$\bullet$]  In Sec.~\ref{sec:qualitative}, we show the detailed qualitative results.

    \item[$\bullet$]  In Sec.~\ref{sec:abl}, we provide ablations about hyperparameters.

    \item[$\bullet$]  In Sec.~\ref{sec:visual}, we provide visualization results for long videos.

    \item[$\bullet$]  In Sec.~\ref{sec:limitation}, we discuss the failure cases, limitations and potential future works.

\end{itemize}

\section{More Analysis of Flow-based T2V Generation for Extracting Motion Representation}
\label{sec:motion_rep}
\begin{figure*}[t]
    \centering
    \includegraphics[width=1\textwidth]{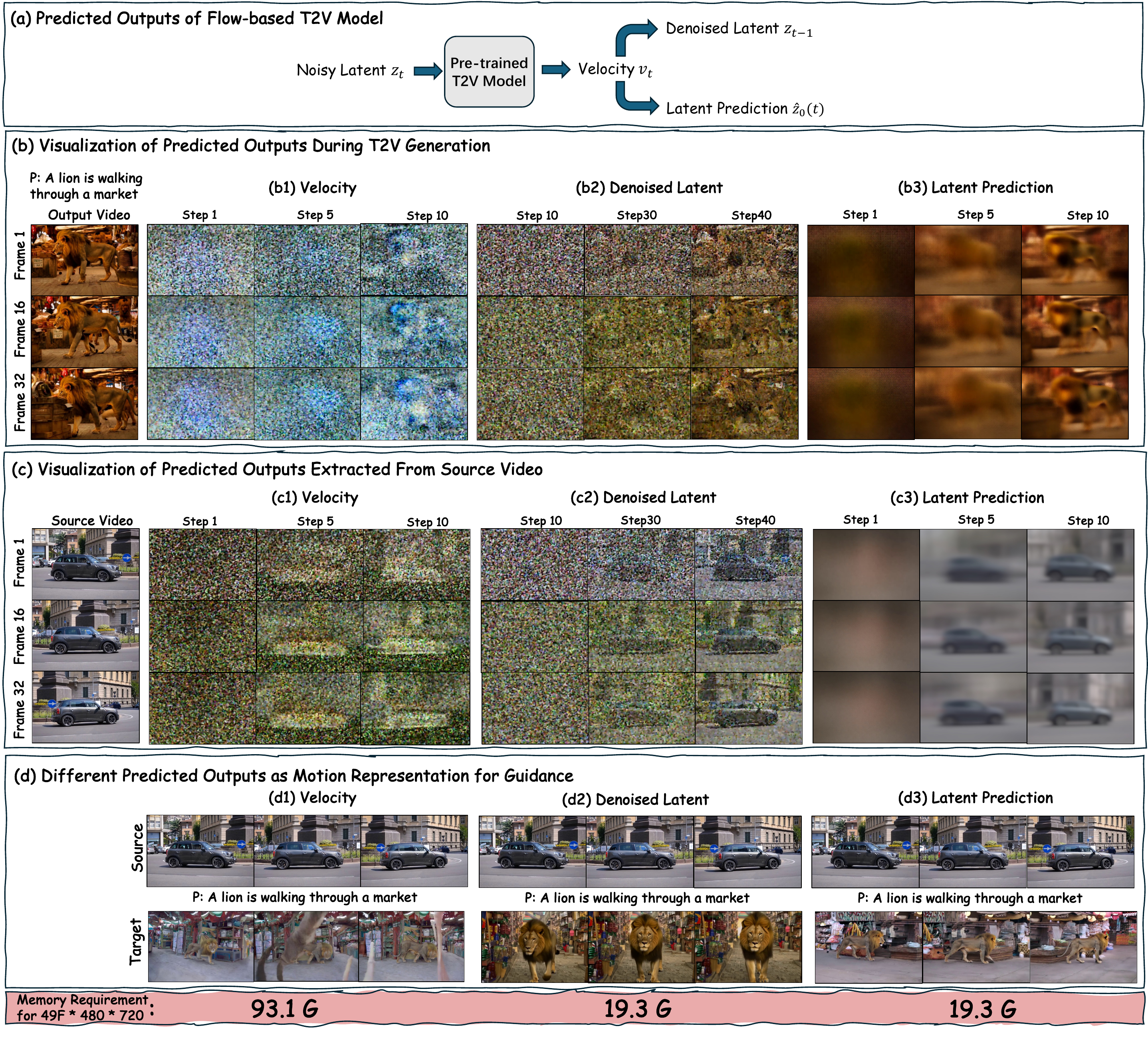}
    \vspace{-2em}
    \caption{Analysis of Flow-based T2V Generation for Extracting Motion Representation.}
    \vspace{-1.5em}
    \label{fig:ana}
\end{figure*}

As stated in Sec.~\textcolor{red}{3.2}, we focus on the model's predicted outputs and investigate whether the intrinsic predictions of flow-based T2V models contain sufficient motion information for direct and efficient transfer. Here, we provide a more detailed analysis of the different predicted outputs and their characteristics. As illustrated in Figure~\ref{fig:ana}(a), at each denoising step of flow-based T2V generation, given a noised latent $z_t$ as input, the model directly predicts the instantaneous \emph{velocity} $v_t$. In the standard denoising process, $v_t$ is used to update $z_t$ with a small step, yielding the \emph{denoised latent} $z_{t-1} = z_t - v_t\, dt$. Furthermore, we introduce a \emph{latent prediction} defined as
$\hat{z}_0(t) = z_t - t \, v_t$, which provides a single-step estimation of the clean latent.

In summary, either directly or indirectly, flow-based T2V models produce three types of predicted outputs: velocity, denoised latent, and latent prediction. In the following, we analyze these outputs through (1) visualizations during the T2V process, (2) extraction and visualization from source videos, and (3) use them as motion representations for guidance. Specifically, we apply the forward noising process described in Sec.~\textcolor{red}{3.2} to source videos to obtain $z^{src}_t$, which are then fed into the T2V model to extract predicted outputs.

\noindent{\textbf{Velocity}.}
As shown in Figure~\ref{fig:ana}(b1), the visualizations of velocity exhibit only coarse and noisy motion trends. Although the motion pattern evolves progressively with the denoising steps, the instantaneous nature of velocity --- representing a noisy-to-clean direction at each step --- makes it inherently noisy throughout generation and difficult to interpret. Similarly, Figure~\ref{fig:ana}(c1) shows that the velocities extracted from source videos remain highly noisy, making the underlying motion signals hard to disentangle. Consequently, as demonstrated in Figure~\ref{fig:ana}(d1), using velocity as the motion representation in our flow guidance leads to imprecise motion transfer: the target video follows only a rough motion pattern and often exhibits abrupt changes. In addition, using velocity for guidance still incurs high memory consumption, as gradient backpropagation through internal layers is required.

\noindent{\textbf{Denoised Latent}.}
As shown in Figure~\ref{fig:ana}(b2), since each denoising step updates only a small portion of the latent, the denoised latents evolve extremely slowly. In the early steps (\eg, the first 10), almost no recognizable motion or appearance information emerges, and meaningful structures only begin to appear at much later steps (around step 30). Moreover, the denoised latents remain highly noisy, and motion and appearance features are heavily entangled, without a clear progression of motion cues. Similarly, Figure~\ref{fig:ana}(c2) shows that denoised latents extracted from source videos exhibit the same characteristics: motion signals are weak, appear only in late steps, and are deeply entangled with appearance information. Consequently, as illustrated in Figure~\ref{fig:ana}(d2), using denoised latent as the motion representation in our flow guidance fails to achieve meaningful motion transfer. The motion cues are too faint and emerge too late for effective optimization, even when the guidance applied throughout the entire denoising process. Thus, although using denoised latent avoids high memory consumption, their guidance effectiveness is severely limited.

\noindent{\textbf{Latent Prediction}.}
From the above analyses of velocity and denoised latent, we can observe that velocity contains highly noisy and hard-to-disentangle motion cues, while the small-step updates in denoised latents provide changes that are too weak and appear only in late denoising stages. These limitations motivate the use of a one-step approximation of the clean latent, \ie, the latent prediction computed directly from velocity as described in the main paper. As shown in Figure~\ref{fig:ana}(b3), early-stage latent predictions inherently encode rich temporal information, exhibiting a clear progressive evolution from coarse spatial locations, to object trajectories, and eventually to fine-grained actions and scene-level dynamics. Correspondingly, Figure~\ref{fig:ana}(c3) shows that latent predictions extracted from source videos follow the same pattern, capturing motion that unfolds progressively from global movement to detailed temporal dynamics. As illustrated in Figure~\ref{fig:ana}(d3), using latent predictions as the motion representation in our flow guidance achieves accurate motion transfer while avoiding the high memory cost. This makes latent prediction both an effective and efficient choice for constructing guidance for motion transfer.

\begin{figure*}[t]
    \centering
    \includegraphics[width=1\textwidth]{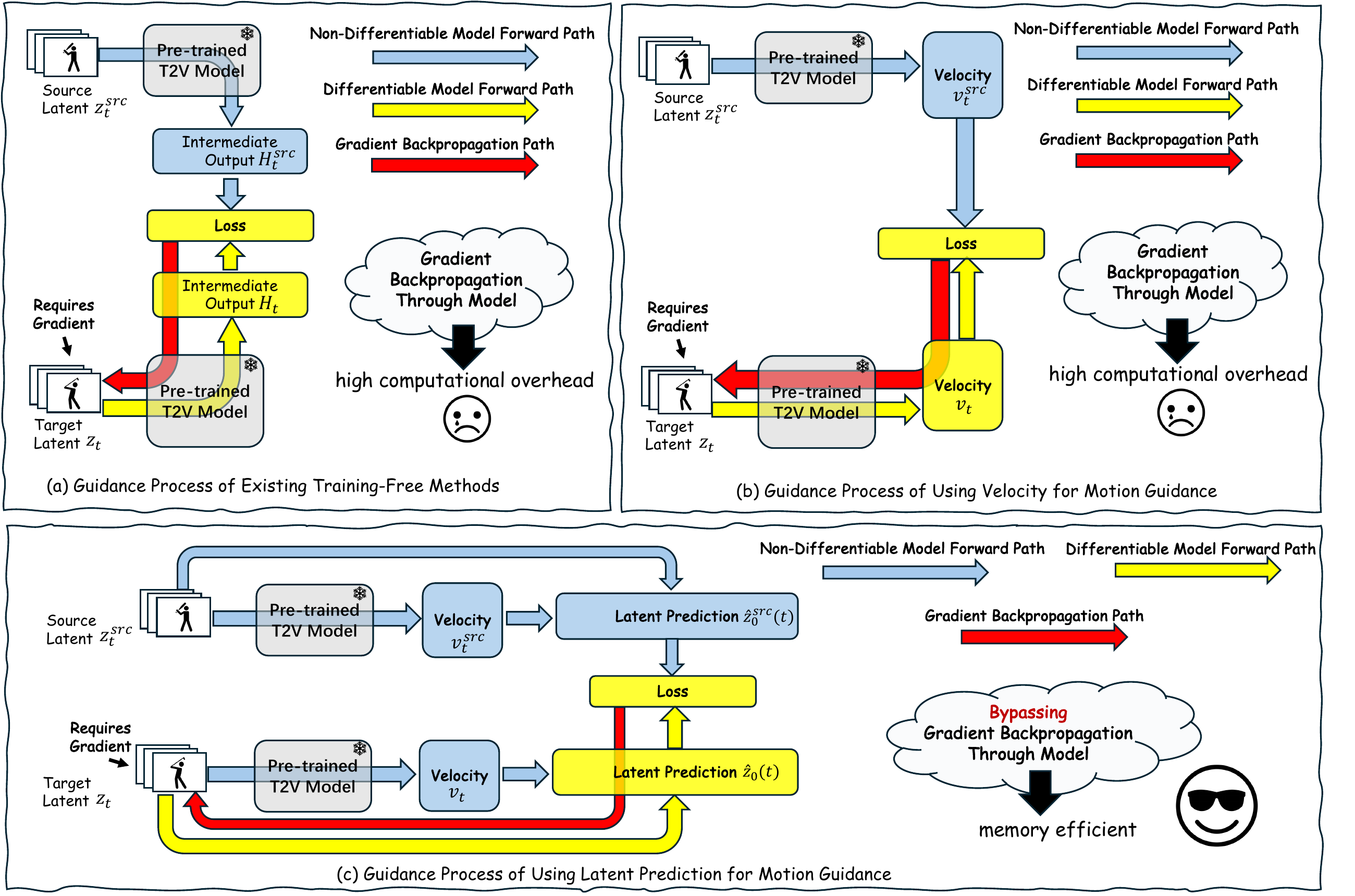}
    \vspace{-2em}
    \caption{Visualization of the Guidance Process of Different Methods.}
    \vspace{-1.5em}
    \label{fig:gradient}
\end{figure*}

\section{Justification for Memory Efficiency}
\label{sec:memory}
We provide a more detailed analysis of the guidance mechanisms used in different methods to clarify their memory consumption. In training-free video motion transfer pipelines, the core principle is to optimize the latent of the target video by minimizing the discrepancy between the motion representations of the source and the generated target video. Consequently, the overall process typically involves two stages: 1) extracting appropriate motion representations from both the source and target videos, and 2) computing a loss between these representations and backpropagating through the model to update the target latent.

\noindent\textbf{Existing Training-Free Methods.}
As shown in Figure~\ref{fig:gradient}(a), existing training-free methods construct motion representations from intermediate model outputs, such as diffusion features or temporal attention maps. For the source video, the intermediate representation $H^{src}_t$ is obtained through a non-differentiable forward pass, which incurs no gradient cost. In contrast, the target latent $z_t$ must undergo a differentiable forward pass to extract its intermediate output $H_t$, based on which a loss is computed against $H^{src}_t$ to update $z_t$ via gradient descent. 

Crucially, since the target latent must be updated by gradient descent, the entire computation from $z_t$ to $H_t$ must remain \emph{fully differentiable}, meaning that all internal model layers involved in producing $H_t$ are retained in the computational graph. As a result, the gradient backpropagation must traverse these deep layers, which further leads to substantial memory consumption.

\noindent\textbf{Using Velocity for Motion Guidance.}
As discussed in Sec.~\ref{sec:motion_rep}, the instantaneous predicted output, \ie, the velocity $v_t$, can also serve as a motion representation. However, as illustrated in Figure~\ref{fig:gradient}(b), obtaining the target velocity still requires a differentiable forward pass from $z_t$ through the flow model. Since the loss between $v_t$ and the source velocity $v^{src}_t$ is used to update $z_t$, this forward pass must remain fully differentiable, and the resulting gradient backpropagation again traverses the internal deep layers. Thus, directly using velocity-based guidance incurs similarly high memory consumption.

\begin{figure*}[t]
    \centering
    \includegraphics[width=1\textwidth]{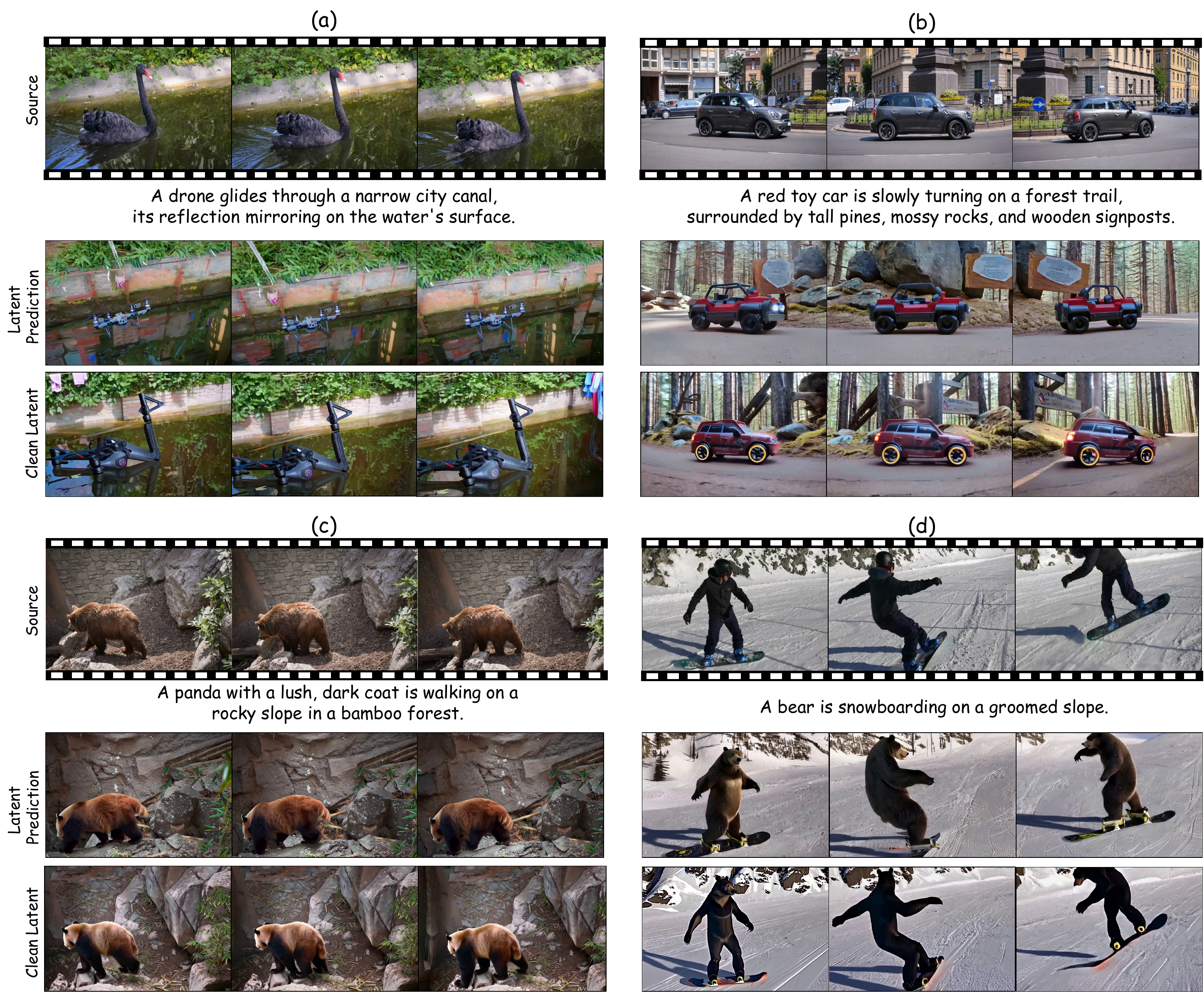}
    \vspace{-2em}
    \caption{Video Motion Transfer Results with Variant of Source Motion Representations.}
    \vspace{-1.5em}
    \label{fig:var}
\end{figure*}

\noindent\textbf{Using Latent Prediction for Motion Guidance.}
As introduced in the main paper and Sec.~\ref{sec:motion_rep}, we define the latent prediction as
$\hat{z}_0(t) = z_t - t\,v_t$, which provides a single-step estimation of the clean latent. As shown in Figure~\ref{fig:gradient}(c), we use these latent predictions as motion representations and compute a loss between the target latent prediction $\hat{z}_0(t)$ and the source latent prediction $\hat{z}^{src}_0(t)$.

A key design lies in how $\hat{z}_0(t)$ is constructed. Although the target velocity $v_t$ is involved in its definition, $\hat{z}_0(t)$ is ultimately a direct function of the target latent $z_t$. Thus, for gradient-based optimization, only the computational graph from $\hat{z}_0(t)$ to $z_t$ needs to be preserved. The velocity $v_t$ can be treated as a constant obtained from a non-differentiable model forward pass, so the forward path producing $v_t$ does not need to be differentiable and is excluded from the computational graph. Consequently, the loss gradient is backpropagated directly through $\hat{z}_0(t)$ to $z_t$, completely bypassing the internal model layers. This eliminates the need to store activations for backpropagation and yields significantly lower memory consumption.

Besides, building on the above design, using the denoised latent $z_{t-1} = z_t - v_t\, dt$ for motion guidance can also reduce memory consumption. However, as discussed in Sec.~\ref{sec:motion_rep}, directly using denoised latent for guidance is less effective in achieving meaningful motion transfer.

The proposed latent prediction offers two key advantages: 1) it operates directly on the model's predicted outputs, without requiring intermediate features such as cross-frame attention that depend on specific model architectures. 2) by virtue of its definition, it elegantly bypasses backpropagation through internal model layers, significantly reducing memory consumption.

In summary, this latent-level flow guidance leverages the model’s predicted outputs, avoids architectural dependencies, and simultaneously achieves both computational efficiency and effective motion transfer.

\section{Variant of Source Motion Representation}
\label{sec:variant}

We provide additional discussion on the variant of source motion representation referenced in Sec.~\textcolor{red}{4.4}. Since the latent prediction can be viewed as a one-step approximation of the final clean latent, thus for the source video,  we can actually directly obtain the ``real'' clean latent, \ie, the clean source latent $z^{src}_0$. This allows a natural variant in which the flow guidance is constructed using $z^{src}_0$ as the source motion representation, while the target motion representation remains the step-wise latent prediction $\hat{z}_0(t)$.

For this variant, we highlight two key differences:  
1) Using latent prediction $\hat{z}^{src}_0(t)$ for the source yields a step-aligned representation that corresponds one-to-one with the target motion representation $\hat{z}_0(t)$. As analyzed and visualized in Sec.~\ref{sec:motion_rep}, both exhibit a similar coarse-to-fine temporal evolution across denoising steps, leading to a more naturally aligned guidance signal.  
2) In contrast, using the source latent $z^{src}_0$ provides a stronger and cleaner motion cue at every step, since it encodes the fully denoised motion information of the source video, but also introduces entangled appearance details that may lead to suboptimal motion alignment. We provide corresponding visualizations in Figure~\ref{fig:var}, where the two variants are respectively used as the source motion representation to construct our flow guidance for motion transfer, enabling a detailed comparison of their behaviors.

As shown in Figure~\ref{fig:var}(a)(b), for normal or easy motions where the dominant dynamics lie in global trajectory changes rather than shape or pose deformations, latent predictions offer a more balanced transfer. They preserve motion fidelity while maintaining text alignment and scene diversity. In contrast, using the clean source latent injects strong appearance information, which may cause appearance leakage into the target object. For instance, in Figure~\ref{fig:var}(a), the generated ``drone" inherits the ``swan"'s outline, and in Figure~\ref{fig:var}(b), the synthesized ``toy car" resembles the source vehicle in shape and appearance, together with reduced background diversity.

Conversely, Figures~\ref{fig:var}(c)(d) illustrate cases with fine-grained or pose-dependent motion such as limb articulation, body deformation, or coordinated multi-part dynamics where the clean source latent performs better. It preserves nuanced pose transitions and detailed actions, accurately transferring the ``bear"’s leg motion in Figure~\ref{fig:var}(c) and the complex human action in Figure~\ref{fig:var}(d). While latent predictions still capture the coarse motion trend, their transfer tends to be more global and less precise at the action level. Nevertheless, latent predictions consistently achieve stronger text alignment and more diverse visual appearance, whereas clean source latent sacrifices these aspects due to stronger entangled cues.

In summary, latent predictions are well suited for most normal or general motions including global trajectory changes or commonly occurring actions, offering a natural balance between motion fidelity and text alignment. Clean source latent, while appearance-entangled, serves as an effective alternative for challenging cases where motion is strongly tied to object pose or fine-grained shape dynamics. Therefore, in this work, we treat the clean source latent as an optional variant specifically for these complex scenarios.

These observations further reveal two promising future directions: 1) Combining latent prediction and clean latent to form a more comprehensive latent-level guidance that simultaneously strengthens fine-grained motion transfer and text alignment; 2) Designing adaptive guidance policies that allow controllable transfer strength, \ie, dynamically selecting which aspects of the motion (global trajectory or detailed actions) to transfer based on the nature of the source video. As the first work to leverage latent-level model outputs for guidance, we hope these findings will stimulate further progress in this emerging direction.

\begin{algorithm}[t]
\caption{\methodname inference pipeline}
\label{alg:motion-transfer}
\begin{algorithmic}[1]
\REQUIRE Source video $\mathcal{V}$, pre-trained flow-based T2V model $v_\theta$, encoder $\mathcal{E}$, decoder $\mathcal{D}$, prompt $P$.
\ENSURE Generated video $x_0$ with transferred motion
\STATE Extract source latent representation: $z^{ref}_0 \gets \mathcal{E}(\mathcal{V})$
\STATE Initialize $z_1 \sim \mathcal{N}(0, I)$
\FOR{denoising step $t_d= 1$ to $T$ with timestep $t = 1$ to $0$}
    \IF{$t_d < T_\text{opt}$}

    \STATE Sample $\epsilon_t \sim \mathcal{N}(0, I)$
    \STATE Forward Noising: $z^{src}_t = (1-t) z^{src}_0 + t \epsilon_t $
    \STATE Compute $\hat{z}^{src}_0(t) = z^{src}_t - t v_\theta(z^{src}_t, t, \emptyset )$
    \STATE Compute $\bigtriangleup(\hat{z}^{src}_0(t))$ 
        \FOR{optimization step $k = 0$ to $K_\text{opt}$}

            \STATE Compute $v_t = v_\theta(z_t, t, P)$

            \STATE Compute $v^{avg}_{t}, v^{proj}_{t}, v^{orth}_t$
            
            \STATE Regulate $v^{reg}_{t} = v^{proj}_{t} + \gamma \cdot v^{orth}_{t}$
            
            \STATE Extract $\hat{z}_0(t) = z_t - t\, v^{reg}_{t}$
            \STATE Compute $\bigtriangleup(\hat{z}_0(t))$
            
            \STATE Get $\mathcal{L}_{FG}  =  \alpha \left \| \hat{z}^{src}_0(t)- \hat{z}_0(t) \right \|^{2}_{2}
 + $
            \STATE $\beta \left \| \bigtriangleup (\hat{z}^{src}_0(t)) - \bigtriangleup (\hat{z}_0(t)) \right \|^{2}_{2}$
            \STATE Update $z_t$ by minimizing $\mathcal{L}_{FG}$
        \ENDFOR
    \ENDIF
    \STATE $z_{t-1} = z_t - v_\theta(z_t, t, P)dt$
    \ENDFOR
 
\RETURN $x_0 = \mathcal{D}(z_0)$
\end{algorithmic}
\end{algorithm}

\section{Inference Pipeline}
\label{sec:algri}
We show the full inference pipeline in Algorithm\ref{alg:motion-transfer}. The whole process built upon a general T2V inference process, start form a random noise $z_1\sim \mathcal{N}(0, I)$, we perform $T$ denoising steps ($t_d= 1$ to $T$) with corresponding timestep $t = 1$ to $0$ to generate the final video $x_0 = \mathcal{D}(z_0)$. Specifically, we apply the guidance in the first $T_\text{opt}$ denoising step for optimization (\ie, $t_d < T_\text{opt}$). Furthermore, for each denoising step we apply optimization, we can optimize the latent for iteratively for $K_\text{opt}$ optimization steps. As discussed in Sec~\ref{sec:variant}, we can also replace $\hat{z}^{src}_0(t)$ with $z^{src}_0$ as a variant.

\begin{table*}[t]
  \centering
 % \vspace{-2mm}

  \resizebox{0.95\textwidth}{!}{%
    % \small  % Increased font size
    \begin{tabular}{l|ccc|ccc|ccc|ccc|ccc}
      \toprule
      \multirow{2}{*}{Method} & \multicolumn{3}{c|}{Easy} & \multicolumn{3}{c|}{Medium} & \multicolumn{3}{c|}{Hard}
        & \multicolumn{3}{c|}{All}
        & \multicolumn{3}{c}{Efficiency} \\
      \cmidrule(lr){2-16} 
        & TS$\uparrow$ & MF$\uparrow$  & TC$\uparrow$ & TS$\uparrow$ & MF$\uparrow$  & TC$\uparrow$ &  TS$\uparrow$ & MF$\uparrow$  & TC$\uparrow$ &TS$\uparrow$ & MF$\uparrow$  & TC$\uparrow$ & 
        Train(s) & Infer(s) & Mem(G) \\
      \midrule
      \multicolumn{9}{l}{\textbf{\emph{Training-based}}} \\
      \midrule
    LoRA Tuning~\cite{hu2022lora} & 0.333 & 0.752 & 0.977 & 0.321 & 0.843 & 0.982 & 0.323 & 0.761 & 0.971
        & 0.327 & 0.782 & 0.977 & 8100
        & 135  & 25.0 \\
    MotionDirector~\cite{zhao2024motiondirector} & 0.338 & 0.767 & 0.973 & 0.333 & 0.895 & 0.972 & 0.337 & 0.752 & 0.962
        & 0.335 & 0.801 & 0.969 & 1662
        &  140 & 28.0 \\
    MotionInversion~\cite{wang2025motioninversion} & 0.327 & \underline{0.802} & 0.973 & 0.325 & \underline{0.909} & 0.973 & 0.334 & \textbf{0.820} & 0.964
        & 0.328 & \underline{0.839} & 0.970 & 1170
        & 115 & 24.0\\
    DeT~\cite{shi2025decouple}  & 0.348 & 0.774 & 0.980 & 0.335  & 0.898 & 0.983 & 0.334 & 0.777 & 0.976
    & 0.340 & {0.812} & 0.980 & 
       2760 & 133 & 20.0 \\
      \midrule
      \multicolumn{9}{l}{\textbf{\emph{Training-free}}} \\
      \midrule
    MotionClone~\cite{ling2024motionclone} &0.332 & 0.788 & 0.947 & 0.332 & 0.836 & 0.943 & 0.333 & 0.734 & 0.929
        & 0.332 & 0.786  & 0.940 & - 
        &  804 & 51.5 \\
    MOFT~\cite{xiao2024video} & 0.346 & 0.631 & 0.972 & 0.326 & 0.599 & 0.979 & \underline{0.341} & 0.501 & 0.970
        & 0.338  & 0.582  & 0.973 & - 
        &  576 & 75.0 \\
    SMM~\cite{yatim2024space} & 0.323 & 0.716 & 0.961 & 0.320 & 0.862 & 0.959 & 0.325 & 0.727 & 0.953
        & 0.322 & 0.762  & 0.958 & - 
        & 1839  & 89.4  \\
    DiTFlow~\cite{pondaven2025video} & \underline{0.350} & 0.692 & \underline{0.986} & \textbf{0.351} & 0.771 & \underline{0.985} & \textbf{0.350} & 0.611 & 0.976
        & \textbf{0.350} & 0.691  &\underline{0.983} &   -
        & 349  & 63.5 \\
      Ours  & \cellcolor{mygray-bg}{\textbf{0.353}} & \cellcolor{mygray-bg}{\textbf{0.834}} & \cellcolor{mygray-bg}{\textbf{0.988}} & \cellcolor{mygray-bg}{\underline{0.345}} & \cellcolor{mygray-bg}{\textbf{0.912}} & \cellcolor{mygray-bg}{\textbf{0.989}} & \cellcolor{mygray-bg}{0.339} & \cellcolor{mygray-bg}{\underline{0.808}} & \cellcolor{mygray-bg}{\textbf{0.980}}
        & \cellcolor{mygray-bg}{\cellcolor{mygray-bg}{\underline{0.347}}
        }& \cellcolor{mygray-bg}{\cellcolor{mygray-bg}{\textbf{0.850}}}  & \cellcolor{mygray-bg}{\cellcolor{mygray-bg}{\textbf{0.986}}} &  \cellcolor{mygray-bg}{-}
        & \cellcolor{mygray-bg}{213}  & \cellcolor{mygray-bg}{19.3} \\
      \bottomrule
    \end{tabular}%
  }
  \vspace{-0.5em}
    \caption{\textbf{Quantitative comparison with SOTA video motion transfer methods}. Best results in \textbf{bold} and second best are \underline{underlined}. } 
      \label{tab:qualitative}
\vspace{-1em}
\end{table*}

\section{Experimental Setup}
\label{sec:experimental}

\noindent{\textbf{Datasets}.}
MTBench~\cite{shi2025decouple} provides high-quality videos with broad motion diversity, covering (1) various subject domains (\eg, humans, animals, and vehicles) and (2) different levels of motion difficulty, categorized into easy, medium, and hard based on trajectory complexity. However, we observe that MTBench has relatively limited camera motion and contains a certain degree of repetitive video patterns. In addition, some of the evaluation prompts are relatively simple. To construct a diverse yet efficient evaluation set, we primarily select videos from MTBench, supplement them with additional camera-motion videos from existing open-source datasets, and further rewrite and filter the prompts. This process results in a curated evaluation set of 50 videos, each paired with three refined prompts, ensuring both diversity and evaluation efficiency.

\noindent{\textbf{Implementation Details}.}  We apply FlowMotion on two pre-trained flow-based T2V models --- Wan2.1-1.3B and Wan2.2-5B~\cite{Wan}. Generally, for both models, we employ $T=50$ denoising steps for T2V generation with a classifier-free guidance of 6, and apply the optimization for the initial $T_\text{opt} =10$ denoising steps. We use the Adam~\cite{kingma2014adam} optimizer with a learning rate of 0.003 for $K_\text{opt} =3$ optimization steps. We set $\alpha:\beta = 4:1$ for the flow guidance loss, and $\gamma = 0.1$ for the velocity regularization. Besides, based on the observation and discussion provided in Sec~\ref{sec:variant}, for qualitative evaluation, we use the latent prediction as source motion representation for easy and medium videos. And use the clean latent as source motion representation for hard videos. We do not employ any additional memory-saving techniques such as CPU offloading.

\section{Detailed Qualitative Results}
\label{sec:qualitative}

We provide more detailed qualitative results in Table~\ref{tab:qualitative} across different motion difficulty levels. We can observe: 1) {FlowMotion} achieves the most balanced performance across all metrics and difficulty levels with superior computational efficiency. 2) The training-free method DitFlow obtains high text similarity but exhibits poor motion fidelity. In contrast, the training-based method MotionInversion tends to overfit, leading to degraded text alignment and reduced temporal consistency. 3) Notably, all methods perform relatively worse on videos categorized as \textit{hard}, which typically involve complex human actions and multiple-object movements. Besides, most approaches achieve better results on \textit{medium} videos than on \textit{easy} ones. Upon inspection, we find that MTBench’s difficulty annotations sometimes place visually simple videos into the medium category, while certain visually complex videos appear in the easy category. This indicates that developing a more principled and reliable motion-difficulty categorization protocol remains an important direction for future work.

\begin{table}[t]
    \centering
    \resizebox{0.48\textwidth}{!}{%
    \begin{tabular}{l|c|c|c}
    \hline
         $\alpha : \beta$ & Text Similarity$\uparrow$ & Motion Fidelity.$\uparrow$  & Temporal Consis. $\uparrow$  \\ \hline
        $1:2$ & 0.336 & 0.742 & \textbf{0.988} \\
       $1:1$ & 0.335 & 0.818 & \textbf{0.988}\\
        $2:1$ & \underline{0.340} & 0.846 & \underline{0.986} \\
       $4:1$ & \cellcolor{mygray-bg}{\textbf{0.341}} & \cellcolor{mygray-bg}{\textbf{0.854}}  & \cellcolor{mygray-bg}{\underline{0.986}} \\
        $8:1$ & 0.338 & \underline{0.848} & 0.985 \\
        \hline
    \end{tabular}
    }
      % \vspace{-1em}
    \caption{{Ablation of weighting coefficients for guidance loss}.} 
    % \vspace{-1em}
\label{tab:abl_loss}
\end{table}

\section{More Ablation Study}
\label{sec:abl}
We provide more ablation studies for the hyperparameter choices of flow guidance. Specifically, we run ablations on Wan2.1-1.3B with a subset of 15 videos.

\noindent{\textbf{Weighting Coefficients for Flow Guidance Loss}.}
The weighting coefficients $\alpha$ and $\beta$ control the balance between latent alignment and frame-wise difference alignment. As shown in Table~\ref{tab:abl_loss}, setting $\beta$ larger than $\alpha$ weakens global alignment and leads to degraded motion quality. Increasing $\alpha$ progressively improves overall alignment, and a reasonable trade-off emerges around $\alpha : \beta = 4 : 1$. This indicates that flow guidance primarily relies on global alignment, while frame-wise differences serve as complementary refinement. The flow guidance is not sensitive to hyperparameters, where ratios within a reasonable range (\eg, $2 : 1$ or $4 : 1$) all provide stable performance. We adopt $4 : 1$ as a general purpose setting.

\noindent{\textbf{Decay Factor for Velocity Regularization}.}
For velocity regularization, we decompose the velocity and attenuate its orthogonal component using a decay factor $\gamma \in [0,1]$. As shown in Table~\ref{tab:abl_decay}, removing regularization ($\gamma = 1$) leads to low text alignment and temporal quality, while enabling decay consistently improves the results. Increasing the decay strength (\eg, $\gamma$ from 0.5 to 0.3) does not yield strictly linear performance changes, instead, the results gradually stabilize toward trade-offs. However, setting $\gamma = 0$ eliminates all orthogonal updates, which overly constrains the optimization dynamics and consequently reduces motion fidelity. Besides, in some cases, visual artifacts caused by unstable optimization may not be fully reflected by quantitative metrics. Therefore, we adopt a relatively strong decay factor of $\gamma = 0.1$ as a robust default setting.

\noindent{\textbf{Denoising Step for Optimization}.} As discussed in Sec.\textcolor{red}{3.2}, the motion information evolving progressively during the early denoising phase (approximately the first 10 steps), after which appearance details become increasingly dominant. As shown in Table\ref{tab:abl_de}, applying guidance for only the first 5 denoising steps results in insufficient motion transfer, while extending it to 15 steps tends to overfit appearance and reduce text similarity. Therefore, we generally optimize over the first 10 denoising steps as a balanced trade-off.

\begin{table}[t]
    \centering
    \resizebox{0.48\textwidth}{!}{%
    \begin{tabular}{l|c|c|c}
    \hline
         $\gamma$ & Text Similarity$\uparrow$ & Motion Fidelity$\uparrow$  & Temporal Consis. $\uparrow$  \\ \hline
        1.0 & 0.316 & 0.844 & 0.974 \\
       0.5 & 0.324 & {\textbf{0.878}} & 0.980\\
        0.3 & {0.334} & \underline{0.871} & {0.981} \\
       0.1 & \cellcolor{mygray-bg}{\textbf{0.341}} & \cellcolor{mygray-bg}{0.854}  & \cellcolor{mygray-bg}{\underline{0.986}} \\
        0.0 & \underline{0.338} & 0.809 & \textbf{0.987} \\
        \hline
    \end{tabular}
    }
      % \vspace{-1em}
    \caption{{Ablation of decay factor for velocity regularization}.} 
    % \vspace{-1em}
\label{tab:abl_decay}
\end{table}

\begin{table}[t]
    \centering
    \resizebox{0.48\textwidth}{!}{%
    \begin{tabular}{c|c|c|c}
    \hline
         $T_\text{opt}$ & Text Similarity$\uparrow$ & Motion Fidelity$\uparrow$  & Temp Consis. $\uparrow$  \\ \hline
        5 & \textbf{0.353} & 0.708 & \underline{0.988} \\
       10 & \cellcolor{mygray-bg}{\underline{0.341}} & \cellcolor{mygray-bg}{\underline{0.854}}  & \cellcolor{mygray-bg}{\textbf{0.986}} \\
        15 & 0.325 & \textbf{0.888} & 0.984\\
        \hline
    \end{tabular}
    }
      % \vspace{-1em}
    \caption{{Ablation of denoising step for optimization}.} 
    % \vspace{-1em}
\label{tab:abl_de}
\end{table}

\begin{table}[t]
    \centering
    \resizebox{0.48\textwidth}{!}{%
    \begin{tabular}{c|c|c|c}
    \hline
         $K_\text{opt}$ & Text Similarity$\uparrow$ & Motion Fidelity$\uparrow$  & Temp Consis. $\uparrow$  \\ \hline
        1 & \textbf{0.356} & 0.549 & \textbf{0.987} \\
       3 & \cellcolor{mygray-bg}{\underline{0.341}} & \cellcolor{mygray-bg}{\underline{0.854}}  & \cellcolor{mygray-bg}{\underline{0.986}} \\
        5 & 0.330 & \textbf{0.897} & 0.985\\
        \hline
    \end{tabular}
    }
      % \vspace{-1em}
    \caption{{Ablation of optimization step}.} 
    % \vspace{-1em}
\label{tab:abl_opt}
\end{table}

\begin{table}[t]
    \centering
    \resizebox{0.48\textwidth}{!}{%
    \begin{tabular}{c|c|c|c}
    \hline
         Motion Rep. & Text Sim.$\uparrow$ & Motion Fid.$\uparrow$  & Temp Consis. $\uparrow$  \\ \hline
        Clean Latent & 0.331 & 0.907 & 0.984 \\
       Latent Prediction & \cellcolor{mygray-bg}{0.341} & \cellcolor{mygray-bg}{0.854}  & \cellcolor{mygray-bg}{0.986} \\
        \hline
    \end{tabular}
    }
      % \vspace{-1em}
    \caption{{Ablation of variant for source motion representation}.} 
    % \vspace{-1em}
\label{tab:abl_var}
\end{table}

\noindent{\textbf{Optimization Step}.} For each denoising step which we apply the guidance, we can iteratively optimize the latent with multiple optimization steps. As shown in Table~\ref{tab:abl_opt}, using only a single optimization step leads to insufficient motion transfer, while using five steps tends to overfit and harm text alignment, in addition to increasing runtime. Therefore, we set the number of optimization step to 3 as a reasonable trade-off.

\noindent{\textbf{Variant of Source Motion Representation}.} We further ablate the choice of source motion representation discussed in Sec.\ref{sec:variant}. As shown in Table\ref{tab:abl_var}, using the clean source latent achieves stronger motion transfer but tends to overfit, which harms text alignment. Thus, it is more appropriate to select the representation based on the motion characteristics. In general, latent prediction is preferable for typical motions, while clean source latent is better suited for complex, pose-dependent motions. As the first work to introduce latent-level guidance for motion transfer, we believe future research can explore dynamic or hybrid guidance strategies to achieve more comprehensive improvements.

\noindent{\textbf{About the ``trade-off”.}} Across all ablations and visual analyses, the text alignment and motion fidelity inherently trade off against each other, which is also the core challenge observed in existing motion transfer methods (\cf, Sec.~\ref{sec:qualitative}). Stronger guidance --- such as more denoising steps, more optimization steps, or using clean source latent --- indeed improves motion fidelity, but also increases the risk of overfitting and reduces text similarity and diversity.

When the overall performance is already strong in terms of quantitative metrics, the \textbf{numerical} gain in motion fidelity from stronger guidance often outweighs the corresponding drop in text similarity, yet the \textbf{visual} outcome may shows the opposite trend, \ie, stronger guidance tends to introduce more noticeable artifacts and appearance degradation. Specifically, we observe that once motion transfer is already visually satisfactory, pushing guidance strength can continue to raise motion fidelity metric, yet often brings little visual benefit or even noticeable degradation due to appearance overfitting. For example, in Fig.~\ref{fig:var}(a), clean latent yields higher motion fidelity score simply because the generated shape becomes closer to the source, but visually the overfitted appearance is clearly undesirable.

In practice, video motion transfer is ultimately a generative task, where users care most about perceptual quality and the ability to create diverse scenes while preserving motion. Therefore, when choosing implementation settings, we tend to place slightly more emphasis on text alignment in practice to achieve a more balanced overall result. At the same time, the long-term goal remains to improve both sides of the trade-off simultaneously.

Besides, adjusting the guidance strength according to different scenarios and motion patterns allows flexible control over the transfer behavior. In this paper, we provide a generally applicable setting based on the above observations and findings.

\begin{figure*}[t]
    \centering
    \includegraphics[width=1\textwidth]{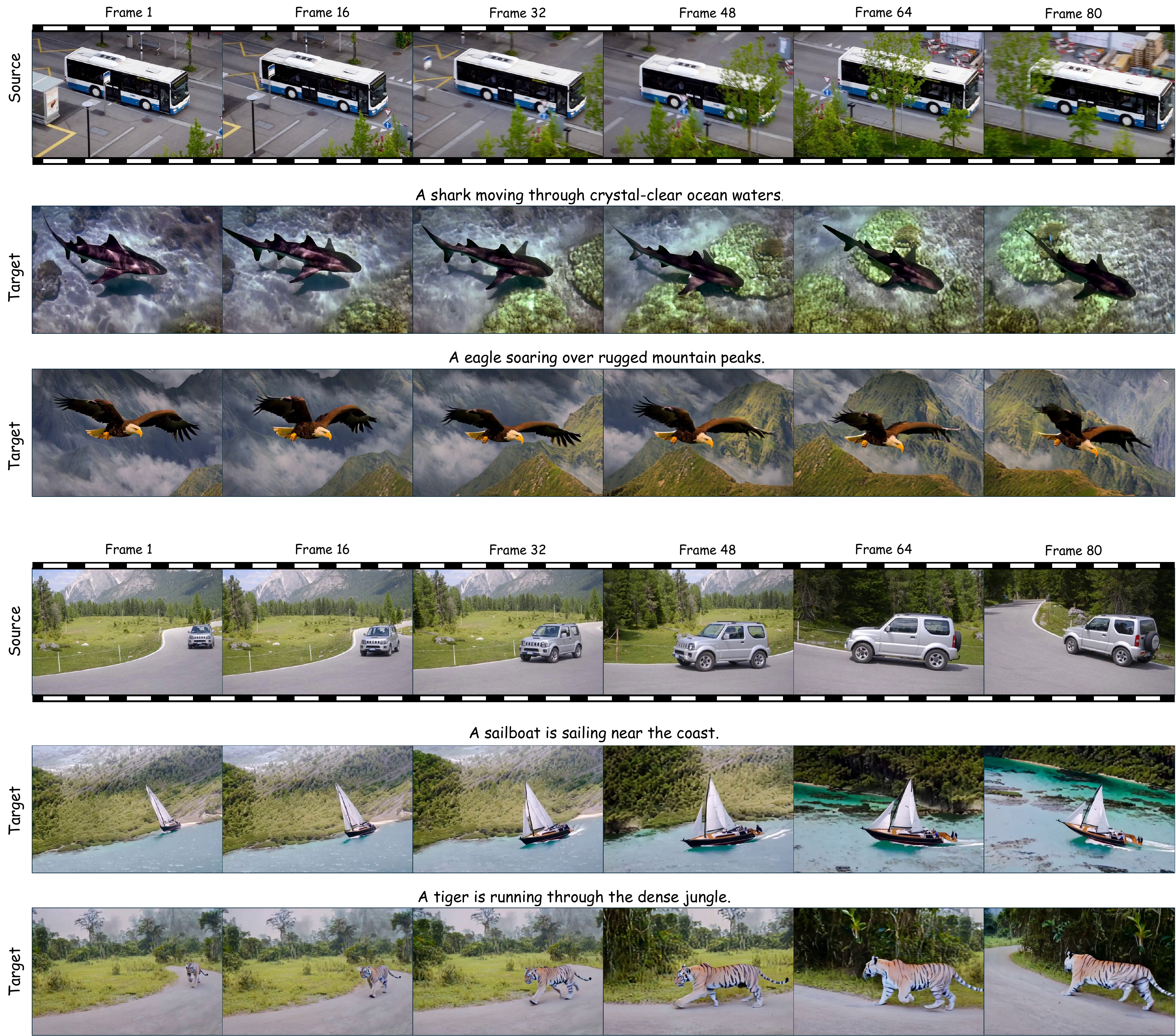}
    % \vspace{-2em}
    \caption{Visualization for video motion transfer on longer videos.}
    % \vspace{-1.5em}
    \label{fig:long}
\end{figure*}

% 20630 7:38

\section{Results for Longer Videos}
\label{sec:visual}
Thanks to the computational efficiency of FlowMotion, our method can be easily applied to longer videos. As shown in Figure~\ref{fig:long}, for videos with 81 frames, FlowMotion achieves effective motion transfer across diverse subjects and scenes. Moreover, for a 480×720 source video with 81 frames, FlowMotion requires only 20.1 G of GPU memory. Since our approach avoids backpropagating through the diffusion model, the additional memory compared with the 49-frame case (19.3G) comes solely from the increased video latent sequence length, resulting in only a marginal GPU overhead. In summary, this efficiency allows our method to run comfortably on consumer-grade GPUs such as the NVIDIA RTX 3090 and 4090, and further underscores FlowMotion’s strong potential for scaling to even longer videos.

\begin{figure*}[t]
    \centering
    \includegraphics[width=1\textwidth]{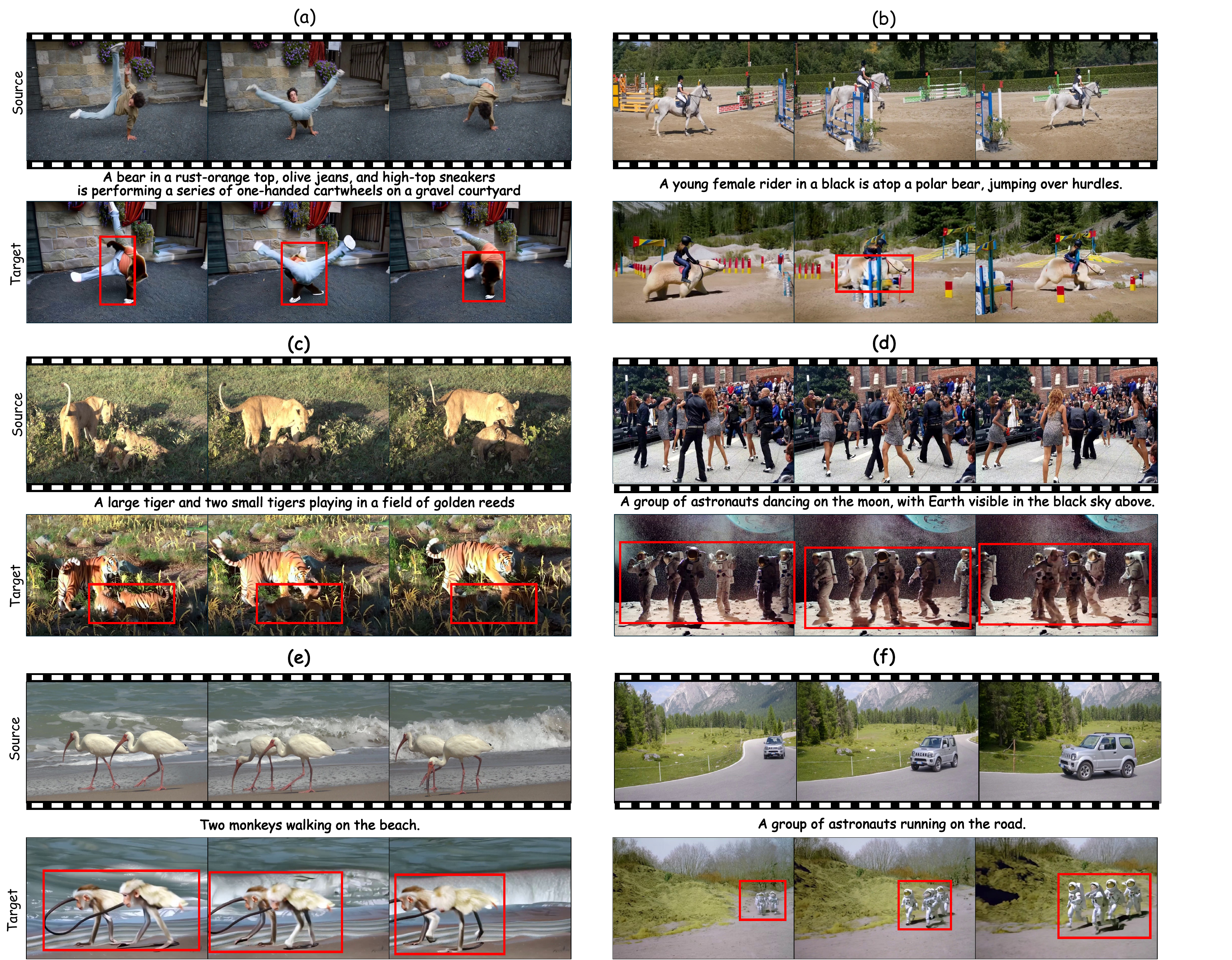}
    % \vspace{-2em}
    \caption{Visualization for the failure cases and limitations.}
    % \vspace{-1.5em}
    \label{fig:lmitation}
\end{figure*}

\section{Limitation and Future Work}
\label{sec:limitation}
We now discuss the failure cases, limitations, and potential future directions. As a training-free framework built on top of a pre-trained T2V model, our method inherently depends on the generation capability of the underlying model. Consequently, it struggles with motion types or prompts that are out of training distribution. In addition, similar to existing methods, as the motion becomes increasingly complex or fine-grained, the transfer ability of the framework gradually reaches its limit.

\noindent{\textbf{Complexity}.} Complex motions involving fine-grained human body movements remain challenging for motion transfer. As shown in Figure~\ref{fig:lmitation}(a), when handling motions such as ``break dance" that involve fine-grained limb coordination, rotations, and uncommon inverted poses, the transferred results exhibit inconsistencies. For example, incorrect hand/foot poses, distorted limbs, or insufficient text alignment of the bear’s appearance. Moreover, for composite motions, as illustrated in Figure~\ref{fig:lmitation}(b), where multiple object movements and camera motion co-occur, the transferred motion may become inaccurate, such as failing to reproduce the ``jumping" motion at the correct location.

As the first work adopting latent prediction as guidance, our approach still operates at a relatively global latent level. Future work could explore incorporating more structured and fine-grained motion cues, such as human keypoints, trajectory tracking, or other explicit motion priors to achieve more precise alignment in complex scenarios.

\noindent{\textbf{Number of Objects}.} As the number of objects increases, and as each object exhibits its own distinct motion, the transfer performance gradually becomes constrained. As shown in Figure~\ref{fig:lmitation}(c), for a scene involving three objects, the transfer for the largest tiger remains relatively accurate, while the two smaller tigers are not well preserved. Furthermore, as illustrated in Figure~\ref{fig:lmitation}(d), when the number of objects exceeds five together with crowd in the background, it becomes increasingly difficult to maintain accurate motion transfer for all objects, and even the object count may become inconsistent.

Notably, existing motion transfer methods primarily focus on single-object scenarios or, at most, two objects. Thus, multi-object motion remains an open challenge. For future work, incorporating spatial priors such as object masks to perform localized latent prediction may enable object-specific motion extraction and improve multi-object motion transfer.

\noindent{\textbf{Transfer Strength and Semantic Gap}.} Existing methods still struggle to automatically determine ``how much" motion should be transferred to the target video. When there is a large semantic gap between the subject described in the target prompt and the subject in the source video, the transfer quality becomes notably limited. As shown in Figure~\ref{fig:lmitation}(e), when transferring the motion of a ``sea birds" to ``monkeys", the model cannot reliably decide whether to retain the global trajectory or the fine-grained pose patterns. This misalignment often leads to visually inconsistent results or distortions, such as generating monkeys that are partially shaped like birds.

Similarly, as illustrated in Figure~\ref{fig:lmitation}(f), transferring the motion of ``a single car" to ``a group of astronauts" introduces semantic mismatch, resulting in degraded performance, such as sudden changes in the number of astronauts or inaccurate motion trajectories for each individual.

Future work may explore adaptive guidance strategies that enable controllable transfer strength. For example, dynamically choosing whether to emphasize global trajectories or detailed actions based on the characteristics of the source motion and the semantics of the target prompt.

\end{document}